\documentclass[10pt,twocolumn,letterpaper]{article}

\usepackage[pagenumbers]{cvpr} % To force page numbers, e.g. for an arXiv version

\usepackage{graphicx}
\usepackage{amsmath}
\usepackage{amssymb}
\usepackage{booktabs}

\usepackage{times}
\usepackage{epsfig}
\usepackage{color}

\usepackage{algorithm}
\usepackage{algorithmic}

\usepackage{multirow}
\usepackage{adjustbox}
\usepackage{bbm}

\usepackage{tablefootnote}
\usepackage[pagebackref,breaklinks,colorlinks]{hyperref}

\usepackage[capitalize]{cleveref}
\crefname{section}{Sec.}{Secs.}
\Crefname{section}{Section}{Sections}
\Crefname{table}{Table}{Tables}
\crefname{table}{Tab.}{Tabs.}

\def\cvprPaperID{9649} % *** Enter the CVPR Paper ID here
\def\confName{CVPR}
\def\confYear{2023}

\begin{document}

% \title{DistractFlow: Improving Optical Flow Estimation Models\\ via Realistic Distractions and Pseudo-Labeling}

\title{DistractFlow: Improving Optical Flow Estimation via\\ Realistic Distractions and Pseudo-Labeling}
% probably no need to mention confidence/uncertianty here, since we don't have a lot of novelty there. -h

% \and
\author{
Jisoo Jeong~~~
Hong Cai~~~
Risheek Garrepalli~~~
Fatih Porikli~~~
\smallskip
\\
Qualcomm AI Research$^{\dagger}$
\\
\smallskip
{\tt\small\{jisojeon, hongcai, rgarrepa, fporikli\}@qti.qualcomm.com}
}
\maketitle

%%%%%%%%% ABSTRACT
\begin{abstract}
%\fp{please check the abstract, I reworded it}
We propose a novel data augmentation approach, DistractFlow, for training optical flow estimation models by introducing realistic distractions to the input frames. 
Based on a mixing ratio, we combine one of the frames in the pair with a distractor image depicting a similar domain, which allows for inducing visual perturbations congruent with natural objects and scenes. We refer to such pairs as distracted pairs. 
Our intuition is that using semantically meaningful distractors enables the model to learn related variations and attain robustness against challenging deviations, compared to conventional augmentation schemes focusing only on low-level aspects and modifications. 
More specifically, in addition to the supervised loss computed between the estimated flow for the original pair and its ground-truth flow, we include a second supervised loss defined between the distracted pair's flow and the original pair's ground-truth flow, weighted with the same mixing ratio. Furthermore, when unlabeled data is available, we extend our augmentation approach to self-supervised settings through pseudo-labeling and cross-consistency regularization. Given an original pair and its distracted version, we enforce the estimated flow on the distracted pair to agree with the flow of the original pair. 
Our approach allows increasing the number of available training pairs significantly without requiring additional annotations. It is agnostic to the model architecture and can be applied to training any optical flow estimation models. 
Our extensive evaluations on multiple benchmarks, including Sintel, KITTI, and SlowFlow, show that DistractFlow improves existing models consistently, outperforming the latest state of the art. % (while halving errors for the most potent models). \js{I think that it's only the case for RAFT on Slowflow (blur duration 3 frames).}
%, and sets the best scores. 

%In this paper, we propose a novel approach, DistractFlow, to augment the training of optical flow estimation models by introducing realistic distractions to the input frames. More specifically, for each pair of training video frames, we combine the second frame with another image based on a mixing ratio. This introduces visual perturbations from real-world objects and scenes; we refer to such a pair as a distracted pair. The prediction on the distracted pair of frames is supervised with the ground truth of the original pair and the loss is added to the original supervised loss. In this way, the model learns to be more robust w.r.t. high-level, semantic variations, as compared to conventional augmentation schemes that focus on low-level aspects of the images. In addition, when unlabeled data is available, by leveraging our novel augmentation of realistic distractions and pseudo-labeling, we can impose a self-supervised regularization. Specifically, given an original pair of frames and its distracted version, we enforce the prediction on the distracted pair to match the prediction on the original pair. This makes the model perform consistently despite visual distractions. Our proposed approach can be applied to the training of any optical flow estimation models. Based on extensive evaluation on optical flow estimation benchmarks like Sintel, KITTI, and SlowFlow, we show that DistractFlow improves existing optical flow estimation models consistently and outperforms the latest state of the art (e.g., reducing error of RAFT by 50\% on SlowFlow).
\end{abstract}

% -------------------------------------------------------
% -------------------------------------------------------
% -------------------------------------------------------
\section{Introduction}
\label{sec:intro}
\vspace{-3pt}

{\let\thefootnote\relax\footnotetext{{
\hspace{-6.5mm} $\dagger$ Qualcomm AI Research is an initiative of Qualcomm Technologies, Inc.}}}

% Optical Flow is a very important key for Video analysis. It is widely used in Frame interpolation, object tracking, Video compression, and so one. So, improvements of optical flow directly benefits other tasks. 

Recent years have seen significant progress in optical flow estimation thanks to the development of deep learning, e.g.,~\cite{dosovitskiy2015flownet, ilg2017flownet, ranjan2017optical, hur2019iterative}. Among the latest works, many focus on developing novel neural network architectures, such as PWC-Net~\cite{sun2018pwc}, RAFT~\cite{teed2020raft}, and FlowFormer~\cite{huang2022flowformer}.
% \cite{ranjan2017optical, jiang2021learning, zhang2021separable}. 
Other studies investigate how to improve different aspects of supervised training~\cite{sun2022disentangling}, e.g., gradient clipping, learning rate, and training compute load. More related to our paper are those incorporating data augmentation during training (e.g.,~\cite{teed2020raft}), including color jittering, random occlusion, cropping, and flipping. While these image manipulations can effectively expand the training data and enhance the robustness of the neural models, they fixate on the low-level aspects of the images.

\begin{figure}[t]
\centering
\includegraphics[width=0.99\linewidth]{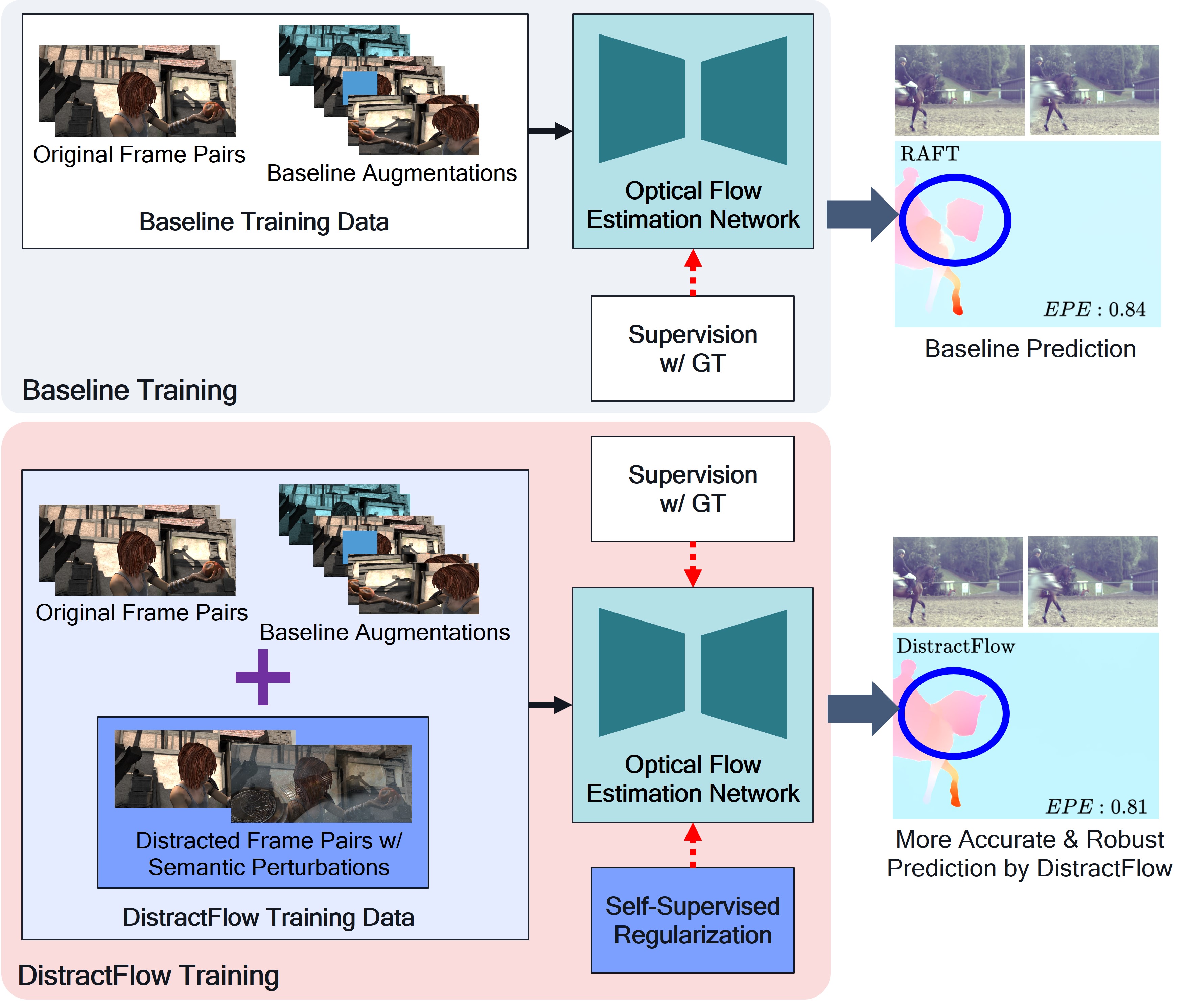}
\vspace{-8pt}
\caption{Existing augmentation schemes apply low-level visual modifications, such as color jittering, block-wise random occlusion, and flipping, to augment the training data (top), while DistractFlow introduces high-level semantic perturbations to the frames (bottom). DistractFlow can further leverage unlabeled data to generate a self-supervised regularization. Our training leads to more accurate and robust optical flow estimation models, especially in challenging real-world settings. 
% \fp{should we remove two "Scheme" in the above figure?}
}
\label{intro:introex}
\vspace{-12pt}
\end{figure}

Since obtaining ground truth optical flow on real data is very challenging, another line of work investigates how to leverage unlabeled data. To this end, semi-supervised methods~\cite{im2022semi, jeong2022imposing} that utilize frame pairs with ground-truth flow annotations in conjunction with unlabeled data in training have been proposed. For instance, FlowSupervisor~\cite{im2022semi} adopts a teacher-student distillation approach to exploit unlabeled data. This method, however, does not consider localized uncertainty but computes the loss for the entire image between the teacher and student network.

In this paper, we present a novel approach, DistractFlow, which performs semantically meaningful data augmentations by introducing images of real objects and natural scenes as distractors or perturbations to training frame pairs. More specifically, given a pair of consecutive frames, we combine the second frame with a random image depicting similar scenarios based on a mixing ratio. In this way, related objects and scenes are overlaid on top of the original second frame; see Figure~\ref{intro:introex} for an example. As a result, we obtain challenging yet appropriate distractions for the optical flow estimation model that seeks dense correspondences from the first frame to the second frame (and in reverse too). The original first frame and the composite second frame constitute a \emph{distracted} pair of frames, which we use as an additional data sample in both supervised and self-supervised training settings. 
Unlike our approach, existing data augmentation schemes for optical flow training apply only low-level variations such as contrast changes, geometric manipulations, random blocks, haze, motion blur, and simple noise and shapes insertions \cite{teed2020raft, sun2021autoflow}.  While such augmentations can still lead to performance improvements, they are disconnected from natural variations, scene context, and semantics. 
% \js{DistractFlow, on the other hand, can blend images to add semantic information of objects from other images. In DistractFlow, object from original image may overlap with the object or background. Even in the case of the background overlapping, there are many noise such as leaves of tree, wall of building, lane on the road, etc, which can have the same effect as original augmentation 
% % Hi Fatih, I got it. Thanks. 
% %hi Jisoo, about the comment on "realistic" intuitions, let's try to add some more explanation in 3.1. I will polish it too. Thanks!
% can add semantic information of object from other images by mixing the images. In DistractFlow, 
% These objects can be not only rigid object but also non-rigid object such as human and tree. }
As we shall see in our experimental validation, the use of realistic distractions in training can provide a bigger boost to performance.

Figure~\ref{intro:introex} provides a high-level outline of DistractFlow. We apply DistractFlow in supervised learning settings using the ground-truth flow of the original pair. Distracted pairs contribute to the backpropagated loss proportional to the mixing ratios used in their construction. Additionally, when unlabeled frame pairs are available, DistractFlow allows us to impose a self-supervised regularization by further leveraging pseudo-labeling. Given an unlabeled pair of frames, we create a distracted version. Then, we enforce the estimated flow on the distracted pair to match that on the original pair. In other words, the prediction of the original pair is treated as a pseudo ground truth flow for the distracted pair. Since the estimation on the original pair can be erroneous, we further derive and impose a confidence map to employ only highly confident pixel-wise flow estimations as the pseudo ground truth. This prevents the model from reinforcing incorrect predictions, leading to a more stable training process.

In summary, our main contributions are as follows:
\vspace{-0.5mm}
\begin{itemize}
\vspace{-0.5mm}
% \item We propose a novel approach, DistractFlow, to augment optical flow training, by utilizing introducing distractions from natural images. As compared to conventional augmentation schemes, our method provides augmentation that contains realistic semantic contents. 
\item We introduce DistractFlow, a novel data augmentation approach that improves optical flow estimation by utilizing distractions from natural images. Our method provides augmentations with realistic semantic contents compared to existing augmentation schemes.

\vspace{-0.5mm}
% \item Utilizing the realistic distractions from DistractFlow and pseudo labeling, we additionally introduce a new semi-supervised learning scheme for optical flow, which allows us to leverage unlabeled data when available. During the process of semi-supervised learning, we further compute a confidence map that provides stability to the training. 
\item We present a semi-supervised learning scheme for optical flow estimation that adopts the proposed distracted pairs to leverage unlabeled data. We compute a confidence map to generate uncertainty-aware pseudo labels and to enhance training stability and overall performance.
% \item We present a semi-supervised learning scheme for optical flow estimation that adopts the proposed distracted pairs to leverage unlabeled data. We compute a confidence map to generate uncertainty-aware pseudo labels and to enhance training stability and overall performance.
%To the best of our knowledge, it is the first attempt to use pseudo-label in semi-supervised optical flow method. \js{applying confidence map for pseudo-labeling is the first attempt. But, there are not many research for semi-supervised optical flow. }

\vspace{-0.5mm}
% \item Through extensive experiments, we demonstrate the effectiveness of our proposed DistractFlow method. When only using labeled data, DistractFlow improves state-of-the-art (SOTA) optical flow estimation models~\cite{teed2020raft, jiang2021learning, huang2022flowformer}, and even outperforms the latest SOTA training scheme of Flow Supervisor~\cite{im2022semi} that requires extra, in-domain unlabeled data. When we leverage additional unlabeled data, DistractFlow then enables further accuracy improvements. 
\item We demonstrate the effectiveness of DistractFlow in supervised~\cite{teed2020raft, jiang2021learning, huang2022flowformer} and semi-supervised settings, showing that DistractFlow outperforms the very recent FlowSupervisor~\cite{im2022semi} that require additional in-domain unlabeled data. %Combined with DistractFlow, further improvement of performance in semi-supervised setting}

\end{itemize}

% -------------------------------------------------------
% -------------------------------------------------------
% -------------------------------------------------------
\section{Related Work}
\label{sec:related}
\vspace{-3pt}
\textbf{Optical Flow Estimation:}
Several deep architectures have been proposed for optical flow \cite{dosovitskiy2015flownet, ilg2017flownet, ranjan2017optical, sun2018pwc, zhao2020maskflownet, teed2020raft}. Among these, Recurrent All Pairs Field Transforms (RAFT) \cite{teed2020raft} have shown significant performance improvement over previous methods, inspiring many subsequent works \cite{jiang2021learning, zhang2021separable, stone2021smurf, huang2022flowformer, sun2022disentangling}. Following the structure of RAFT architecture, complementary studies \cite{xu2021high, jeong2022imposing, jiang2021learning, zhang2021separable, zhao2022global} proposed advancements on feature extraction, 4D correlation volume, recurrent update blocks, and more recently, transformer extensions \cite{zhao2022global, huang2022flowformer}. In DistractFlow, we introduce a new model-agnostic training method that can help any model.

\begin{figure*}[ht]
\centering
\includegraphics[width=0.95\linewidth]{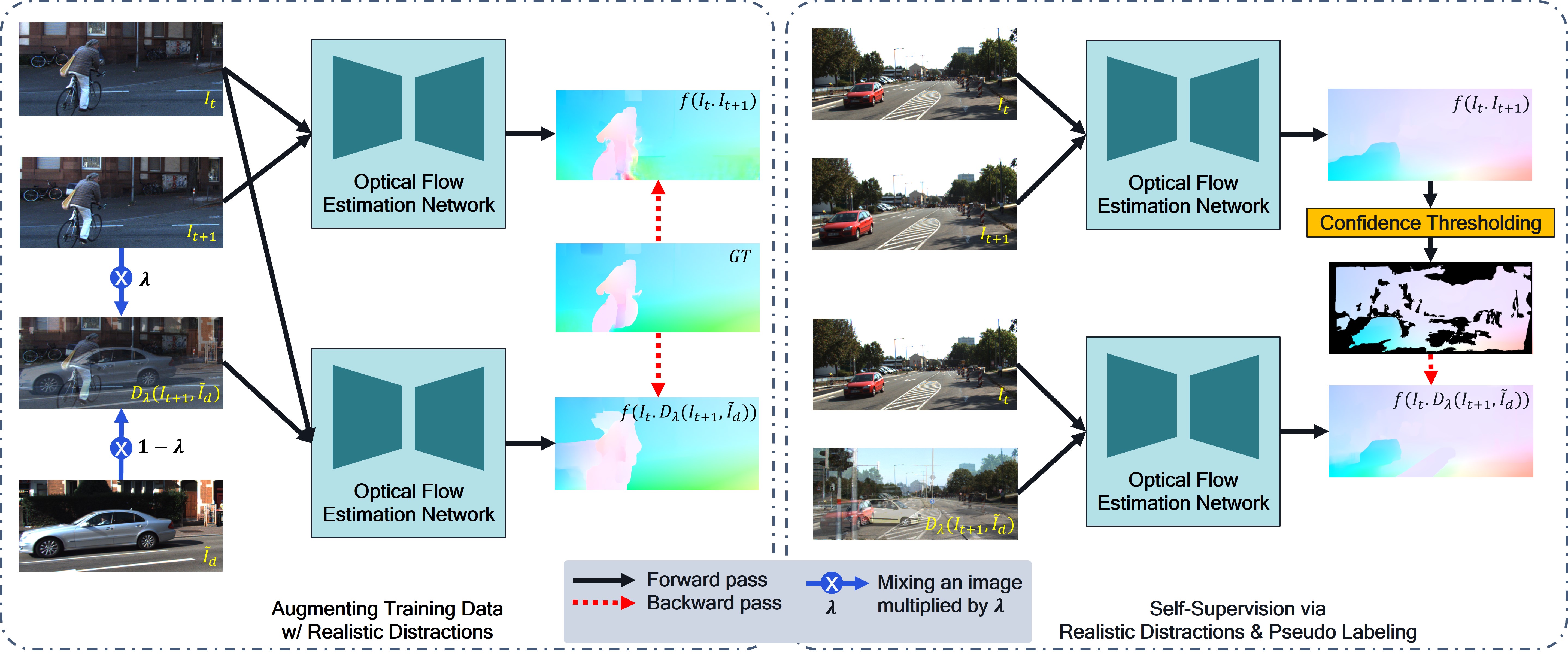}
\vspace{-10pt}
\caption{Left: Introducing realistic distractions in the supervised learning setting. Right: Semi-supervised learning leveraging distractions and pseudo-labeling. }
\vspace{-10pt}
\label{fig:method_sup}
% \vspace{-5mm}
\end{figure*}

\textbf{Data Augmentation:}
Data augmentation is a widely used technique to better train deep learning models. Common augmentations include color and contrast jittering, flipping, geometric manipulation, and random noise. While they can improve the robustness of the model, these operations mainly focus on the low-level visual aspects and do not account for variations in the semantic contents.

Recently, several data augmentation schemes, such as adversarial perturbation and regularization methods, have been proposed for classification tasks. Adversarial perturbation~\cite{xie2020adversarial} is one of the well-known augmentation methods for classification tasks, but recent work~\cite{schrodi2022towards} shows that it does not work well for optical flow estimation. 
Interpolation-based Regularization (IR)~\cite{zhang2017mixup, yun2019cutmix}, which mixes a couple of images and trains the model with mixed label, improves the performance in classification tasks and is employed in other fields such as object detection \cite{zhang2019bag, jeong2021interpolation} and segmentation \cite{islam2022segmix}. 
In a regression problem, however, mixed ground truth does not correspond with mixed input.
Fixmatch~\cite{sohn2020fixmatch} as a data augmentation scheme has demonstrated state-of-the-art performance in classification tasks. It combines pseudo-labeling and consistency regularization by applying two different augmentations (weak, strong) to the same image and generates pseudo-labels with weak augmentation output. For classification tasks, it is possible to create pseudo-labels since the output and ground-truth annotations represent class probabilities. However, there are no such class probabilities for optical flow. Thus, these methods cannot be readily applicable.

% . In a regression problem, however
% In addition, Adversarial perturbation 
% }

% Fixmatch \cite{sohn2020fixmatch} as data augmentation scheme demonstrated state-of-the-art performance in classification tasks. It combines pseudo-labeling and consistency regularization by applying two different augmentations (weak, strong) to the same image and generates pseudo-labels with weak augmentation output. If the output of the weak augmentation is over a threshold, it trains the model with the pseudo-label and the strong augmented output. For classification tasks, it is possible to create pseudo-labels since the output and ground-truth annotations represent class probabilities. On the other hand, there are no such class probabilities for optical flow. Thus, Fixmatch cannot be readily applicable. 

% \hc{[AutoFlow. Are there other more advanced augmentations?]}
% \rg{It might be important to adopt generalizable architectures like RAFT and good augmentation and regularization strategies with diverse training towards robustness in OOD/Openset and adversarial settings \cite{vaze2022openset,pmlr-v80-liu18e,liu2022pac}}.
AutoFlow \cite{sun2021autoflow} proposed a new dataset for optical flow, taking a versatile approach to data rendering, where motion, shape, and appearance are controlled via learnable hyperparameters. Though its performance gain is notable, AutoFlow employs synthetic augmentations. The work in \cite{sun2022disentangling} utilizes AutoFlow and argues that it is important to disentangle architecture and training pipeline. 
\cite{sun2022disentangling} also points out that some of the performance improvements of the recent methods are due to hyperparameters, dataset extensions, and training optimizations. 
Our work focuses on more capable augmentations for model agnostic training, not architecture novelties.

\textbf{Semi-Supervised Learning:}
Semi-supervised optical flow learning methods \cite{lai2017semi, jeong2022imposing} aim to make the best use of unlabeled data as there is only a limited amount of optical flow ground truth data for real and natural scenes. And even in those datasets, the optical flow ground truths are computed. RAFT-OCTC \cite{jeong2022imposing} proposed transformation consistency for semi-supervised learning, which applies spatial transformations to image pairs and enforces flow equivariance between the original and transformed pairs. 
% RealFlow \cite{han2022realflow} generated a new training dataset based on a pretrained model and updated the network iteratively to attain performance gains. \js{shall we just remove RealFlow? since they train the model with unlabeled sintel and kitti train dataset and validate on sintel and train dataset. And, they don't share the sintel leaderboard number. (our model shows better number on Sintel setting, but not kitti)}
FlowSupervisor~\cite{im2022semi} introduced a teacher network for stable semi-supervised fine-tuning. Its student model is trained for all pixels using teacher network output. In DistractFlow, we propose uncertainty-aware pseudo-labeling, which uses two different image pairs instead of different networks. We further employ forward-backward consistency to derive dense confidence scores, which steer the training process to impose loss only within high-consistency image regions to prevent feedback from incorrect flow estimates.

Since acquiring optical flow ground truth for real videos is problematic (not possible for most cases), unsupervised training methods \cite{meister2018unflow, jonschkowski2020matters, stone2021smurf} seek out training models without ground truth flows. Even though they report promising results comparable to the earlier deep learning approaches, unsupervised training methods still entail limitations.

\section{DistractFlow}
\label{sec:method}\vspace{-3pt}

Our approach incorporates augmentation and supervision techniques to enhance the training of optical flow estimation models. In Section~\ref{sec:method_sup}, we describe how we construct realistic distractions for optical flow training and how we employ them in supervised settings. Next, in Section~\ref{sec:method_self}, we extend our approach to semi-supervised learning with additional unlabeled data. We derive a self-supervised regularization objective by utilizing the distracted samples and pseudo-labeling. 
% In Section.\ref{sec:method_total}, we discuss how to implement DistractFlow efficiently in training from a practical perspective. 
%In this section, we present our proposed approach, DistractFlow, which contains novel augmentation and supervision techniques to enhance the training of optical flow estimation models. We first describe how to apply realistic distractions to optical flow training samples as well as how to supervise them in Section~\ref{sec:method_sup}. Next, we consider the semi-supervised learning case, where additional unlabeled data is available. We derive a self-supervised regularization by utilizing the distracted samples and pseudo-labeling in Section~\ref{sec:method_self}. In Section~\ref{sec:method_total}, we discuss how to efficiently implement our proposed schemes in training from a practical angle. 

% \subsection{Interpolation Regularization for Optical Flow}
\subsection{Realistic Distractions as Augmentation}\label{sec:method_sup} \vspace{-3pt}

Consider a pair of video frames during training: $(I_t,\, I_{t+1})$. The distracted version of them is denoted as $(I_t,\, D_{\lambda}(I_{t+1}, \tilde{I}_d)$, where $D_{\lambda}(I_{t+1}, \tilde{I}_d)$ is the perturbed second frame obtained by combing with another image $\tilde{I}_d$ based on a mixing ratio of $\lambda \in (0,\,1)$. Specifically, $D_{\lambda}(I_{t+1}, \tilde{I}_d)$ is calculated as $\lambda \cdot I_{t+1} + (1-\lambda) \cdot \tilde{I}_d$, where $\lambda$ is sampled from a Beta($\alpha, \alpha$) distribution, same as defined in \cite{zhang2017mixup, yun2019cutmix}.

Figure~\ref{fig:method_sup} shows an example of a distracted pair of video frames. It can be seen that the actual objects and the real scene from one image are overlaid onto the second one. Such perturbations can reflect challenging real-world scenarios, e.g., foreground/background objects that only start to appear in the second frame, drastic motion blur, out-of-focus artifacts, reflections on specular surfaces, partial occlusions, etc. Furthermore, since the distractions are from the same dataset, the visual context of the original pair and distractor image are similar. For instance, the original pair and distractor depict similar classes (road, car, building, etc.). The spatial arrangement of the class and object regions in those images are similar, e.g., roads are within the lower part of the images, and so do the sky, buildings, and vehicles. While one may attempt to render such scenarios synthetically, our DistractFlow provides a convenient and automatic way that can still capture such natural and semantically related variations. Compared to conventional augmentation methods that use noise or simple shapes, our method allows the model to be more robust to perturbations caused by real-world image contents. 

Additionally, we note that applying realistic distractions to the first or both frames is possible. As we shall see in the experiments, all of these options will result in improved accuracy.

To provide supervision on the distracted pair of frames, we use the ground-truth flow of the original pair. The loss is computed as follows:
\vspace{-1mm}
\begin{equation}\label{eq:sup_loss} \vspace{-1mm}
\mathcal{L}_\text{dist} = \|V^{f}_{(I_t,\, I_{t+1})} - f(I_{t}, D_{\lambda}(I_{t+1}, \tilde{I}_d))\|_{1},
% \vspace{-1mm}
\end{equation}
where $V^f_{(I_t,\, I_{t+1})}$ is the ground-truth forward flow for the original pair $(I_t,\, I_{t+1})$ and $f(\cdot,\, \cdot)$ denotes the predicted flow based on model $f$.

In the supervised learning setting, where all training samples are labeled, the total training loss is then given as follows:
\vspace{-1.5mm}
\begin{equation} \label{eq:total_sup_loss} \vspace{-1mm}
\mathcal{L}_\text{sup} = \mathcal{L}_\text{base} + w_\text{dist}\mathcal{L}_\text{dist},
\end{equation}
where $L_{base}$ is the conventional supervised loss and $w_\text{dist}>0$ weights $\mathcal{L}_\text{dist}$. For iterative models like RAFT, we compute and apply this loss at each recurrent iteration.

\begin{figure}[t]
\centering
\includegraphics[width=0.96\linewidth]{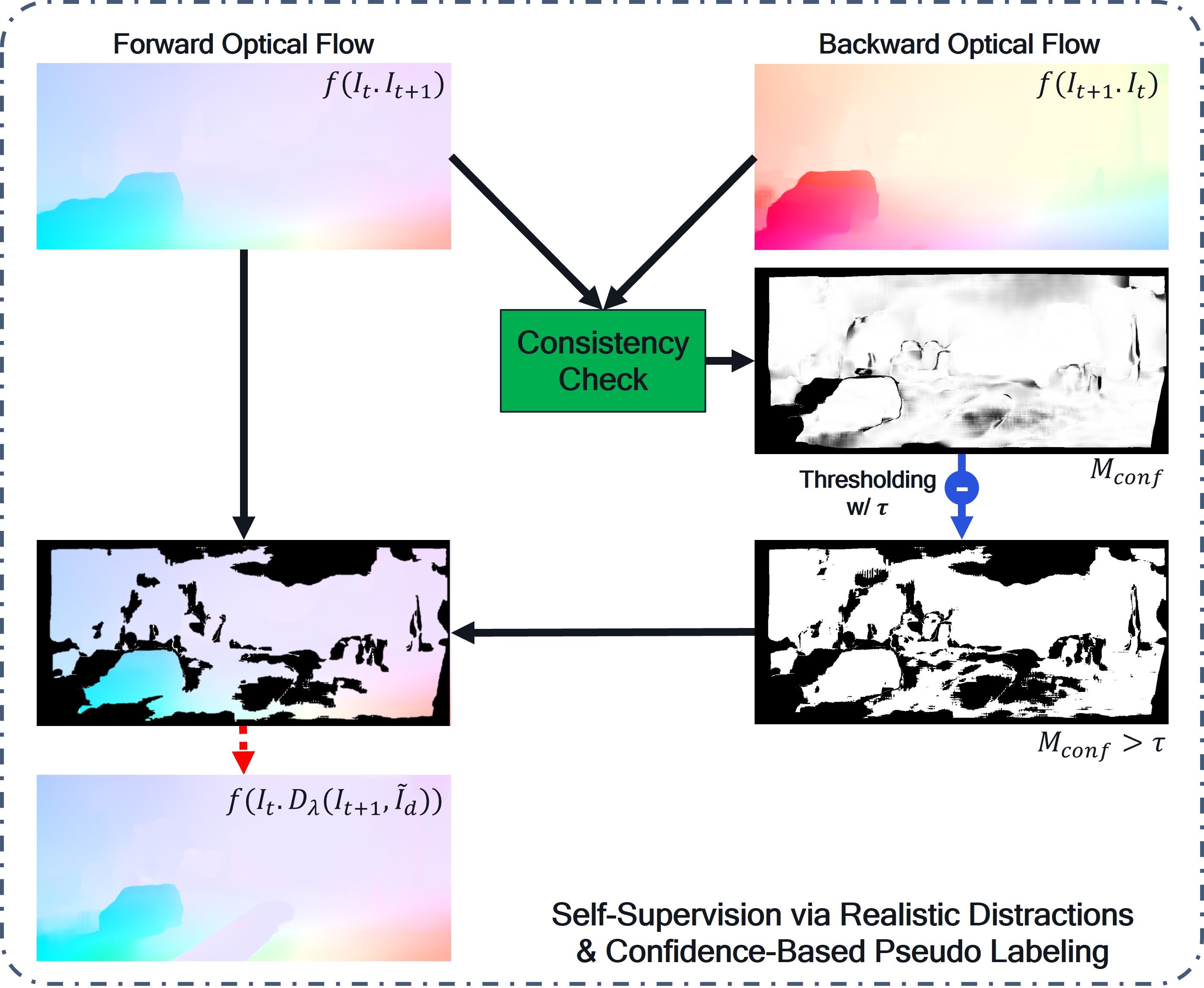}
\vspace{-5pt}
\caption{Semi-supervised learning that leverages distractions and confidence-aware pseudo-labeling.
% \fp{can we enlarge the flow/confidence maps in the figure?}
}
\label{method:semi}
\vspace{-10pt}
\end{figure}

\subsection{Semi-Supervised Learning via Realistic Distraction and Pseudo-Labeling}\label{sec:method_self}
\vspace{-3pt}
During training, when unlabeled frame pairs are available, we can further leverage our distracted pairs of frames to derive additional self-supervised regularization. More specifically, given a distracted pair of frames, $(I_t,\, D_\lambda(I_{t+1},\,\tilde{I}_d))$, and the original pair, $(I_t, I_{t+1})$, we enforce the model's prediction on the distracted pair to match that on the original pair. In other words, the prediction on the original pair, $f(I_t,\, I_{t+1})$, is treated as the pseudo label. By doing this, the model learns to produce optical flow estimation on the distracted pair that is consistent with that of the original pair, despite the distractions. Such regularization promotes the model's robustness when processing real-world data, as we show in our experiments.

We note that, however, using all of $f(I_t,\, I_{t+1})$ as the training target for the distracted pair is problematic. This is because the model's prediction can be erroneous and noisy during the training process, even on the original frame pairs. The low-quality pseudo labels can be detrimental to the overall training, even leading to instability. 

% To address this issue, we calculate a confidence map and only use the highly confident pixel-wise predictions as the pseudo ground truth as shown in Fig.~\ref{method:semi}. 
To address this issue, we adopt uncertainty-aware pseudo labels by calculating a confidence map based on forward backward consistency, and only using highly confident pixels' predictions as pseudo ground truth, as shown in Figure~\ref{method:semi}.
On a frame pair $(I_t,\,I_{t+1})$, let $\widehat{V}^f(x)$ and $\widehat{V}^b(x)$ denote the predicted forward and backward flows at the pixel location $x$. When they satisfy the following constraint~\cite{meister2018unflow}, we can assume that the prediction is accurate. 
% \js{Actually, it is for occlusion map. If it is not satisfied, \cite{meister2017unflow} assume that there is occlusion. Based on this equation, we assume that }
\begin{equation} \label{For-Back_Con}
\begin{split}
% \scriptsize
|\widehat{V}^{f}(x) + \widehat{V}^{b}(x+\widehat{V}^f(x))|^2 \quad \quad \quad \quad \quad \quad \quad \\
< \gamma_{1} \Big{(}|\widehat{V}^{f}|^2 + |\widehat{V}^{b}(x+\widehat{V}^f(x))|^2 \Big{)} + \gamma_{2},
\end{split}
\end{equation}
where $\gamma_1$ = 0.01 and $\gamma_2$ = 0.5 from~\cite{meister2018unflow}.
% \hc{Is this true?}. Yes

As such, we derive the confidence map as follows: 
% \begin{equation} \label{For-Back_Con}
% \begin{split}
% |V^{f}(x) + V^{b}(x+V^f(x))|^2 \quad \quad \quad \quad \quad \quad \quad \\
% < \alpha_{1} \Big{(}|V^{f}|^2 + |V^{b}(x+V^f(x))|^2 \Big{)} + \alpha_{2}
% \end{split}
% \end{equation}
% The pixel is occluded if it violates the constraint \footnote{$\alpha_1$ = 0.01 and $\alpha_2$ = 0.5 are set}.  
\begin{equation} \label{Con_map}
% \begin{split}
M_\text{conf} = \text{exp}  {\Bigg{(}-\frac{|\widehat{V}^{f}(x) + \widehat{V}^{b}(x+\widehat{V}^f(x))|^2}{\gamma_{1} \Big{(}|\widehat{V}^{f}|^2 + |\widehat{V}^{b}(x+\widehat{V}^f(x))|^2 \Big{)} + \gamma_{2} } \Bigg{)}}.
% \end{split}
\end{equation}

Our confidence map 
% quantifies the accuracy 
provides a measure of reliability of the predicted optical flow. Specifically, in Eq.~\ref{Con_map}, if the numerator and the denominator are equal, the confidence value is then approximately 0.37 ($e^{-1}$). In this paper, we use a very high threshold and it could provide a more accurate optical flow pseudo ground truth. 
% \fp{how did we determine 0.37?}
% \js{it is exp(-1). }

%  (over 0.9)\hc{over 0.9 in the linear scale?}
% Then, we compute the unsupervised loss using weak augmented output, strong augmented output, and confidence map $M_{conf}$ in Eq.~\ref{eq:semi_loss1}.
% \begin{equation} \label{Con_map}
% \begin{split}
% l_{unsup} = \mathbb{E}_{\mathbb{I}{\{M\geq \tau\}}} [l_{self}]
% \end{split}
% \end{equation}

\begin{table*}[t]
\begin{center}
\caption{Optical flow estimation results on SlowFlow, Sintel (train), and KITTI (train) datasets. We train the models on FlyingChairs (C) and FlyingThings (T) in the supervised setting. In the semi-supervised setting, we finetune the model on FlyingThings (T) as labeled data and Sintel (test) and KITTI (test) (S/K) as unlabeled data.  
* indicates test results of existing models generated by us. 
% \tablefootnote{Since some scores have large gap between the paper and official code, we report the reproduced number.} 
}
\vspace{-3mm}
\label{tab:supandsemi}
\adjustbox{max width=0.95\textwidth}
{
\begin{tabular}{|l|l|c||c|c||c|c|c|c||c|c|c|}
\hline
% Method & Labeled & Network & \multicolumn{2}{|c|}{mAP (\%)}\\
 & \multirow{2}{*}{Method} &\multirow{2}{*}{Model}&  Labeled & Unlabeled & \multicolumn{4}{|c||}{SlowFlow (100px)}   & Sintel (train)  & \multicolumn{2}{|c|}{KITTI (train)} \\
\cline{6-12}
 &  & & data & data &1&3&5&7 & (Final-epe) & (Fl-epe) & (Fl-all)\\
\hline
\hline
\multirow{6}{*}{\rotatebox{90}{\textbf{Supervised}}}&  Supervised  &  \multirow{2}{*}{RAFT \cite{teed2020raft}} & \multirow{6}{*}{C+T} &  & 3.73 & 7.98 & 6.72 & 9.96 &  $2.73^*$ & $4.94^*$ & $16.9^*$\\
% & GMA \cite{jiang2021learning} & & \textbf{1.30} & 2.74 & 4.69 & 17.1 & - & -  & - \\
& DistractFlow (Our) & & &  &\textbf{3.37} & \textbf{3.93} & \textbf{6.21} & \textbf{8.18} & \textbf{2.61}  & \textbf{4.57} & \textbf{16.4} \\
% \hline
\cline{2-3}
\cline{6-12}
% 1.30 2.74 4.69 17.1
& Supervise &\multirow{2}{*}{GMA \cite{jiang2021learning}} & &   & \textbf{2.49} & 4.99 & 5.95 & 9.15 &  $2.85^*$ & $4.88^*$ & $17.1^*$ \\
& DistractFlow (Our) & & & & 2.53 & \textbf{3.49} & \textbf{5.43} & \textbf{8.24} &  \textbf{2.66}  & \textbf{4.76}  & \textbf{16.9} \\
\cline{2-3}
\cline{6-12}
  & Supervised & \multirow{2}{*}{FlowFormer \cite{huang2022flowformer}} & &  & 2.72 & 3.73 & 5.24 & 6.78 & $2.39^*$ & $4.10^*$ & $14.7^*$ \\
& DistractFlow (Our) &  & & & \textbf{2.51} & \textbf{2.77} & \textbf{4.39} & \textbf{6.30} & \textbf{2.31}  & \textbf{4.00}  & \textbf{13.9} \\
\hline
\hline
\multirow{9}{*}{\rotatebox{90}{\textbf{Semi-Supervised}}} &  Supervised & \multirow{5}{*}{RAFT \cite{teed2020raft}} & C + T &  &3.73 & 7.98 & 6.72 & 9.96 &  $2.73^*$ & $4.94^*$ & $16.9^*$\\
\cline{4-4}
& RAFT-A \cite{sun2021autoflow} & & AutoFlow \cite{sun2021autoflow}  &  & - & - & - & - & 2.57 & 4.23 &  -  \\
\cline{4-4}
 & RAFT-OCTC \cite{jeong2022imposing} & & \multirow{8}{*}{C + T} & T (subsampled) & - & - & - & - & 2.67 & 4.72 & 16.3  \\
 & Fixed Teacher \cite{im2022semi} & &  & $S$/$K$ & - & - & - & - & 2.58 & 4.91 & 15.9  \\
& FlowSupervisor \cite{im2022semi} & & & $S$/$K$ & - & - & - & - & 2.46 & 3.35 & \textbf{11.1}  \\
&  DistractFlow (Our) & & &$S$/$K$ & \textbf{2.46} & \textbf{3.60} & \textbf{5.15} & \textbf{6.95} & \textbf{2.35} & \textbf{3.01} & 11.7  \\
\cline{2-3}
\cline{5-12}
  & Supervised & \multirow{2}{*}{GMA \cite{jiang2021learning}} & &  & 2.49 & 4.99 & 5.95 & 9.15 &  $2.85^*$ & $4.88^*$ & $17.1^*$ \\
& DistractFlow (Our) & & & $S$/$K$ & \textbf{2.44} & \textbf{2.79} & \textbf{4.38} & \textbf{6.57} & \textbf{2.31} & \textbf{3.21} & \textbf{11.0}  \\
\cline{2-3}
\cline{5-12}
& Supervised & \multirow{2}{*}{FlowFormer \cite{huang2022flowformer}} & &  &   2.72 & 3.73 & 5.24 & 6.78 & $2.39^*$ & $4.29^*$ & $15.4^*$ \\
 & DistractFlow (Our) & && $S$/$K$ & \textbf{2.48} & \textbf{2.69} & \textbf{4.31} & \textbf{6.29}  & \textbf{2.33} & \textbf{3.03} & \textbf{11.8}  \\
\hline
\end{tabular}
}
\vspace{-7mm}
\end{center}
\end{table*}

By incorporating the confidence map, our self-supervised regularization is then given as follows:
\begin{equation} \label{eq:semi_1}
\mathcal{L}_\text{self} =\|[M_\text{conf} \geq \tau]\ (f(I_{t},\, I_{t+1}) - f(I_{t}, D_{\lambda}(I_{t+1},\, \tilde{I}_d)))\|_{1}, \\
\end{equation}
where $[\cdot]$ is the Iverson bracket and $\tau$ is the confidence threshold.\footnote{By $[M_\text{conf} \geq \tau]$, we note that the Iverson bracket is applied pixel-wise to produce a binary confidence map.} 

In summary, in the semi-supervised learning setting, where labeled and unlabeled data are both used in training, the total loss is as follows:
\vspace{-1mm}
\begin{equation} \label{eq:total_semi_loss} \vspace{-1mm}
\mathcal{L}_\text{total} = \mathcal{L}_\text{sup} + w_\text{self} \mathcal{L}_\text{self},
% \mathcal{L}_\text{total} = \mathcal{L}_\text{base} + w_\text{dist} \mathcal{L}_\text{dist} + w_\text{self} \mathcal{L}_\text{self},
\end{equation}
where $\mathcal{L}_\text{sup}$ is the supervised loss of Eq.~\ref{eq:total_sup_loss}, $\mathcal{L}_\text{self}$ is the self-supervised loss of Eq.~\ref{eq:semi_1}, and $w_\text{self}>0$ weights $\mathcal{L}_\text{self}$.

% \subsection{Overall Loss for Training}\label{sec:method_total}
% The overall loss for training the optical flow estimation model is given as follows:
% \begin{equation} \label{eq:total loss}
% \begin{split}
% % \mathcal{L}_\text{total} = \mathcal{L}_\text{base} + \mathcal{L}_\text{mix} + \mathcal{L}_\text{self}
% \mathcal{L}_\text{total} = w_{base}\mathcal{L}_\text{base} + w_{mix} \mathcal{L}_\text{mix} + w_{self} \mathcal{L}_\text{self}
% \end{split}
% \end{equation}
% We apply this loss to the optical flow algorithms. We set all $w$s to 1 for our experiments. For example, we compute each total loss from each GRU iteration and apply Eq.~\ref{eq:total loss}.

% \subsection{Efficient Training Implementation}
% \hc{Worth mentioning??} \rg{supplementary?}

% -------------------------------------------------------
% -------------------------------------------------------
% -------------------------------------------------------

\section{Experiments}\label{sec:exp}
\vspace{-3pt}
In this section (and in the supplementary), we present a comprehensive evaluation of DistractFlow on several benchmark datasets, compare it with baselines and the latest state-of-the-art (SOTA) methods, and conduct extensive ablation studies. We focus our evaluations on realistic or real-world data, including SlowFlow, Sintel (final), and KITTI; we provide results on Sintel (clean) in the Supplementary.

\subsection{Experimental Setup}
\vspace{-3pt}
\textbf{Evaluation Settings:}
We consider two settings for running the experiments. In the first setting, we consider a supervised learning setting where the network is trained on fully labeled data. The second setting considers semi-supervised learning, where unlabeled data can be used during training, in addition to labeled training data.

\textbf{Evaluation Metrics:} We use the common evaluation metrics for optical flow estimation, including End-Point Error (EPE) and Fl-all, which is the percentage of optical flow with EPE larger than 3 pixels or over $5\%$ of the ground truth. The goal is to lower both of these metrics. 

% the percentage of outliers (Fl-All) averaged over all ground truth pixels .
% Fl-All (%) metric which refers to the percentage of optical flow vectors whose end-point error islarger than 3 pixels or over 5% of ground truth.

\textbf{Datasets:}
Following commonly adopted evaluation protocols in the literature~\cite{teed2020raft, jiang2021learning, zhang2021separable, huang2022flowformer}, we train our model on FlyingChairs (C)~\cite{dosovitskiy2015flownet} and FlyingThings3D (T)~\cite{mayer2016large} for supervised training when evaluating on SlowFlow~\cite{Janai2017SlowFlow} (100px flow magnitude) dataset and the training splits of Sintel (S)~\cite{butler2012naturalistic} and KITTI (K)~\cite{geiger2013vision, menze2015object}. In particular, on SlowFlow, we use 4 blur durations with a larger duration having larger motion blurs. When we evaluate on Sintel test set, we use FlyingThings3D, Sintel training set, KITTI training set, and HD1K (H)~\cite{kondermann2016hci} for supervised training. And, we finetune on KITTI training dataset for evaluation on KITTI test set.
% And, we also evaluate our model on SlowFlow~\cite{Janai2017SlowFlow} dataset. 

For the additional unlabeled data used during training, we followed the same setting from \cite{im2022semi}. We use FlyingThings as labeled dataset and Sintel test set as unlabeled dataset for evaluating on Sintel and SlowFlow, and the video frames from KITTI test raw sequences for evaluating on KITTI training dataset. When we evaluate on Sintel and KITTI test sets, we use the labeled dataset from supervised training settings, and use additional KITTI train raw sequences, Sintel training data (only using every other frame), Monkaa~(M)~\cite{mayer2016large}, and Driving~(D)~\cite{mayer2016large} as the unlabeled datasets. 
% For the best performance of the model, we use not only C+T+S+K+H dataset but also M+D+V dataset as label data and all others as unlabeled data. 
Note that we do not use Sintel and KITTI test sets as unlabeled datasets for Sintel and KITTI test evaluation. We further experiment with utilizing unlabeled open-source Blender videos as additional unsupervised training data, such as Sintel and Big Buck Bunny.

% , and Virtual KITTI2~\cite{cabon2020virtual}

% we use the video frames from KITTI, Monkaa~(M) \cite{mayer2016large}, and Driving (D)~\cite{mayer2016large}. Virtual KITTI2~\cite{cabon2020virtual} and Slow Flow~\cite{Janai2017SlowFlow} are also used as evaluation datasets.
% Following common adopted evaluation protocols in the literature \cite{teed2020raft, jiang2021learning, zhang2021separable, huang2022flowformer}, we use FlyingChairs (C)~\cite{dosovitskiy2015flownet} and FlyingThings3D (T)~\cite{mayer2016large} for supervised training, when evaluating on the training splits of Sintel (S)~\cite{butler2012naturalistic} and KITTI (K)~\cite{geiger2013vision, menze2015object}. When we evaluate on Sintel and KITTI test sets, we use FlyingThings3D, Sintel training set, KITTI training set, and HD1K (H)~\cite{kondermann2016hci} for supervised training. In the setting where additional unlabeled data is used during training, we use the video frames from KITTI, Monkaa~(M) \cite{mayer2016large}, and Driving (D)~\cite{mayer2016large}. Virtual KITTI2~\cite{cabon2020virtual} and Slow Flow~\cite{Janai2017SlowFlow} are also used as evaluation datasets.

\textbf{Networks and Training:}
We use RAFT \cite{teed2020raft}, GMA \cite{jiang2021learning}, FlowFormer \cite{huang2022flowformer} as our baselines, and utilize their official codes.\footnote{RAFT: \url{https://github.com/princeton-vl/RAFT}, GMA: \url{https://github.com/zacjiang/GMA}, FlowFormer: 
% \url{https://github.com/feihuzhang/SeparableFlow} \\ 
\url{https://github.com/drinkingcoder/FlowFormer-Official}} For objectiveness, we train all the baselines in the same framework and reported the results we obtained. We set ($w_{dist}$, $w_{self}$) and $\tau$ to ($\lambda$ (mixing ratio), 1) and 0.95, respectively. For supervised training, we follow RAFT and GMA learning parameters such as optimizer, number of GRU iterations, training iterations, and so on. For semi-supervised training, we use labeled and unlabeled data with a 1:1 ratio. For FlowFormer, we reduce the total batch size and only finetune it due to GPU memory overflow. 
All settings and training details are provided in the Supplementary.
% The total batch size is 8 for RAFT and GMA and it is 2 for Flowformer.  
% However, because of lack of GPU memory, we only finetuned the Flowformer with 2 batch size with already pre-trained model. For semi-supervised setting, we used 4 labeled and 4 unlabeled pairs in each batch, and trained the model with Flyingthings setting. 

% We followed the
% same batch sizes, optimizer, number of GRU iterations,
% and so on.
% For RAFT and GMA, we follow the training st
% , Separable-Flow \cite{zhang2021separable}

% let's think about some kind of visualization to provide insights
% \subsection{Visualization}

% \begin{table}[t]
% \centering
% \caption{Optical Flow Results for SlowFlow dataset (100px Flow Magnitude). We trained the FlyingChairs and FlyingThings followed RAFT. }
% \label{tab:slowflow}
% \adjustbox{max width=0.48\textwidth}
% {
% \begin{tabular}{|l||c|c|c|c|}
% \hline
% \multirow{2}{*}{Method} & \multicolumn{4}{|c|}{Blur Duration (Frames)}\\
% \cline{2-5}
% & 1 & 3 & 5 & 7 \\
% \hline
% RAFT \cite{teed2020raft} & 3.73 & 7.98 & 6.72 & 9.96 \\
% Our (RAFT) & 3.37 & 3.93 & 6.21 & 8.18 \\
% \cline{2-5}
% GMA \cite{jiang2021learning} & 2.49 & 4.99 & 5.95 & 9.15 \\
% Our (GMA) & 2.53 & 3.49 & 5.43 & 8.24 \\
% \hline
% \end{tabular}
% }
% \end{table}

\subsection{Evaluations on SlowFlow, Sintel (train), and KITTI (train)}
\label{exp:sup}\vspace{-3pt}

\textbf{Supervised Setting:}
Table~\ref{tab:supandsemi} shows the performance evaluation of the supervised and semi-supervised training results on SlowFlow, Sintel (train), and KITTI (train). In the top section of Table~\ref{tab:supandsemi}, we provide results of existing supervised models and our models trained using DistractFlow supervised augmentation. Our supervised models trained on FlyingChairs and FlyingThings3D show significant improvements across the test datasets for all the architectures. For SlowFlow with a 3-frame blur duration, the existing RAFT model shows very larges error (over 100 EPE) on a few samples. DistractFlow, on the other hand, leads to more robust results and a much smaller EPE. 

\textbf{Semi-Supervised Setting:}
In this part, we evaluate our proposed approach when additional unlabeled data becomes available. In the bottom section of Table~\ref{tab:supandsemi}, we show results of the previous semi-supervised learning methods and our proposed method. Following FlowSupervisor, we finetune each model on FlyingThings (labeled) and Sintel test (unlabeled), and evaluate on the SlowFlow and Sintel train dataset. For evaluation on KITTI train dataset, we finetune the pre-trained models (from Sintel unlabeled) on FlyingThings (labeled) and raw KITTI test (unlabeled). We can see that DistractFlow (RAFT) shows better performance as compared to RAFT-OCTC, RAFT-A, Fix-Teacher, and FlowSupervisor. Furthermore, we apply our semi-supervised approach to GMA and FlowFormer, and DistractFlow improves their performance.

% \textbf{Evaluation on Sintel and KITTI:}
% Table \ref{tab:supandsemi} shows the performances of the supervised and semi-supervised training results for optical flow. In top of Table \ref{tab:supandsemi}, we provide the results of previous methods and our proposed method. 
% Our supervised model trained on C+T shows the significant improvement on Sintel (final) and KITTI results for all the methods. And, our model trained on C+T+S+K+H also shows the improvement, and this number is already outperformed the Flow-Supervision (FS)~\cite{im2022semi} without additional unlabeled data. 

\begin{table}[t]
\centering
\caption{Optical flow estimation results on Sintel (test) and KITTI (test) datasets. We train the models on FlyingChairs (C), FlyingThings (T), Sintel (S), KITTI (K), and HD1K (H) in the supervised setting. When using the semi-supervised setting/using additional data for training, RAFT-A trains on A+S+K+H+T datasets where A stands for AutoFlow, FlowSupervisor trains on  C+T+S+K+H (labeled) and uses Sintel, KITTI, and Spring as unlabeled datasets. In our case, we use C+T+S+K+H as labeled data and Sintel, KITTI, and Sceneflow (Monkaa, Driving) as unlabeled data. * indicates results obtained using warm-start \cite{teed2020raft}. }
% A→SKHTV
\vspace{-3mm}
\label{tab:testtest}
\adjustbox{max width=0.45\textwidth}
{
\begin{tabular}{|l|c||c|c|}
\hline

\multirow{2}{*}{Method} & \multirow{2}{*}{Model}   & Sintel (test) & KITTI (test)\\
% \cline{2-3}
\cline{3-4}
&  & (Final-epe) & (Fl-all)\\
\hline
\hline
\multicolumn{4}{|c|}{Supervised}\\
\hline
Supervised & \multirow{2}{*}{RAFT \cite{teed2020raft}}  & $2.86^{*}$ & 5.10 \\
DistractFlow (Our) & &  $\textbf{2.77}^{*}$ &  \textbf{4.82} \\
% \cline{2-3}
\hline
\hline
\multicolumn{4}{|c|}{Semi-Supervised / Additional dataset}\\
\hline
Supervised & \multirow{5}{*}{RAFT \cite{teed2020raft}}  & $2.86^*$ & 5.10 \\
RAFT-A \cite{sun2021autoflow}&  &  3.14 &  4.78 \\
RAFT-OCTC \cite{jeong2022imposing} &   & 3.09 & 4.72 \\
FlowSupervisor \cite{im2022semi} &   & $2.79^*$ & 4.85 \\
DistractFlow (Our) & &  $\textbf{2.71}^*$ & \textbf{4.71} \\
\hline
% Supervised & \multirow{2}{*}{FlowFormer\cite{huang2022flowformer}}  & 2.09 & 4.68 \\
% DistractFlow (Our) &   & - & \textbf{4.52} \\
% \hline
\end{tabular}
}
\vspace{-12pt}
\end{table}

\subsection{Evaluations on Sintel (test) and KITTI (test)}
\label{exp:semi} \vspace{-3pt}
\textbf{Supervised Setting:}
The top section of Table~\ref{tab:testtest} summarizes the supervised and semi-supervised training results on Sintel (test) and KITTI (test) datasets. 
From the first part of Table~\ref{tab:testtest}, we can see that our proposed DistractFlow permits significant improvements over the baseline RAFT model. It is noteworthy that without using additional unlabeled data, our model trained under the supervised setting already outperforms the semi-supervised FlowSupervisor, which leverages additional data.

\textbf{Semi-Supervised Setting:}
In the second part of Table~\ref{tab:testtest}, we provide results for semi-supervised methods or methods that train with additional datasets. RAFT-A is trained with an additional AutoFlow dataset but still underperforms the RAFT trained with our DistractFlow approach. 
RAFT-OCTC applies a semi-supervised method as well as changes the architecture for occlusion prediction. Although RAFT-OCTC uses a slightly more complex architecture, our DistractFlow-trained RAFT still shows better performance. In addition, our method also considerably outperforms FlowSupervisor. 
% \js{Furthermore, we apply our semi-supervised method to FlowFormer and DistractFlow also performs better than supervised-only Flow Former.
% % (Depending on the results, I will add. / I expect to get the number at 5pm.) 
% }
% \rg{Furthermore, when we adopt our semi-supervised method to FlowFormer and it improves performance of FlowFormer compared to supervised-only method.\textcolor{red}{jisoo, can you please check if statement is correct}}

% Even if the architecture of RAFT-OCTC is slightly increased, our model still shows better performance. 
% case, it slightly change the architecture, and it applies
% However, our semi-supervised training model shows better performance. And, Our DistractFlow considerably outperforms Flow-Supervisor.
% existing state-of-the-art semi-supervised method of FlowSupervisor.
% Since RAFT-A trained the RAFT model with AutoFlow dataset, our model only shows the improvement on the Sintel dataset. However, our model with an unlabeled additional dataset shows higher performance on Sintel and KITTI.

% \textbf{Semi-Supervise :}
% In bottom of Table~\ref{tab:testtest}, 
% Our proposed DistractFlow considerably outperforms the existing state-of-the-art semi-supervised method of Flow Supervisor.
% \js{Depending on the result of Flowformer, we can add or remove. (Since FlowFormer requires large memory, I am training the model with 2 batch size.  (label:1/unlabel:1) and C+T+S+K+H : 290k pairs, and original training iteration was 120k. So, I trained only sintel, but it shows very very bad. }

\begin{figure*}[t]
\begin{center}$
\centering
\begin{tabular}{c c c c}
% \multicolumn{2}{|c|}{KITTI (train)}
% \multirow{2}{*}{Method}
% \multirow{2}{*}{Image Overlay} & \multirow{2}{*}{GT} & \multicolumn{2}{c}{Baseline} & \multicolumn{2}{c}{DistractFlow}\\
% Image Overlay \/ GT & Baseline & DistractFlow\\
\\
\hspace{-0.2cm} \includegraphics[width=4.2cm]{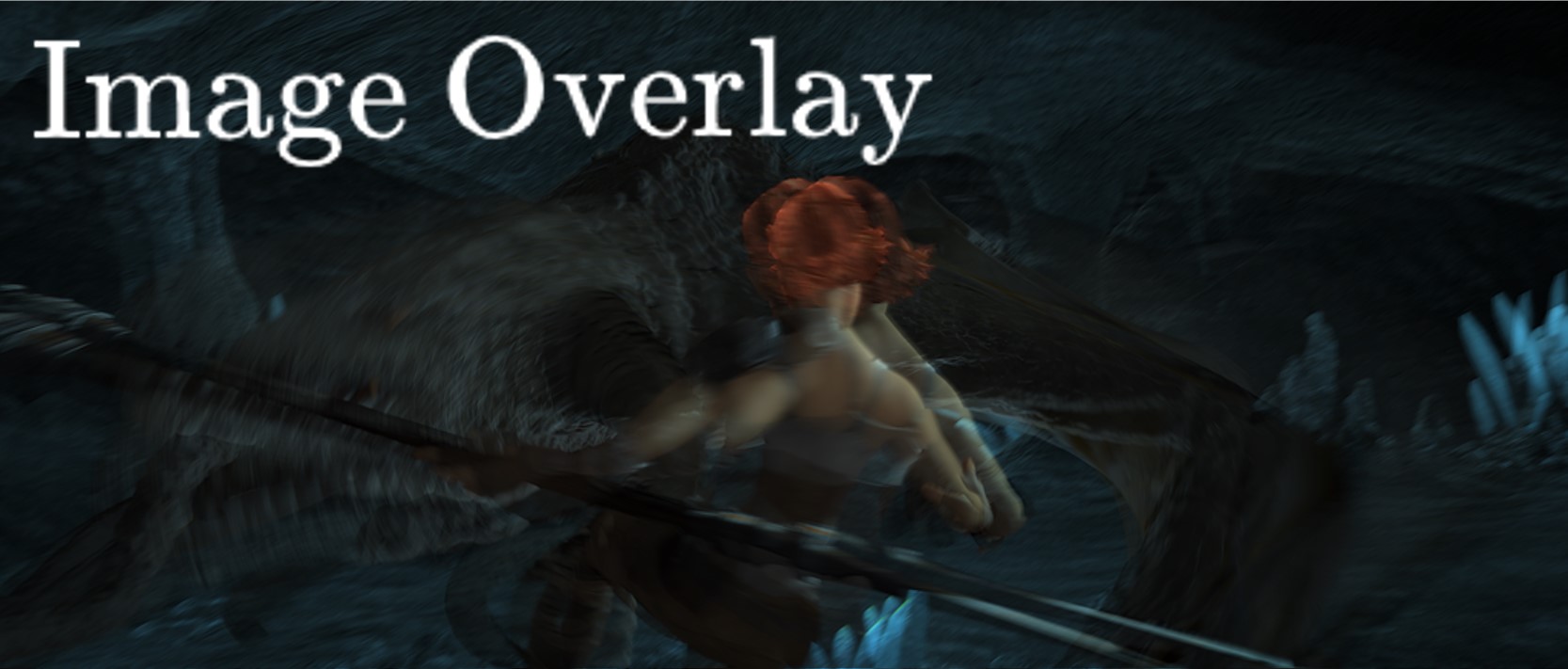}
& \hspace{-0.4cm} \includegraphics[width=4.2cm]{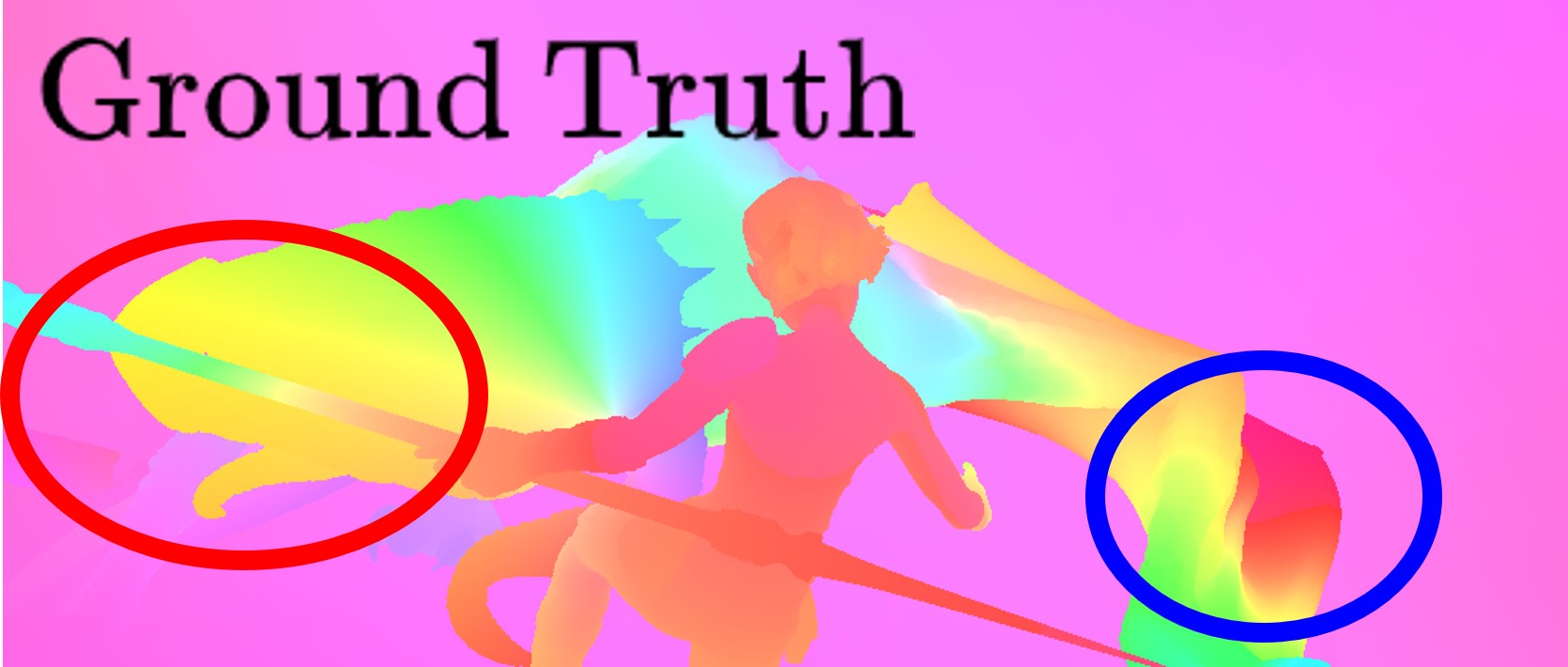}
& \hspace{-0.4cm} \includegraphics[width=4.2cm]{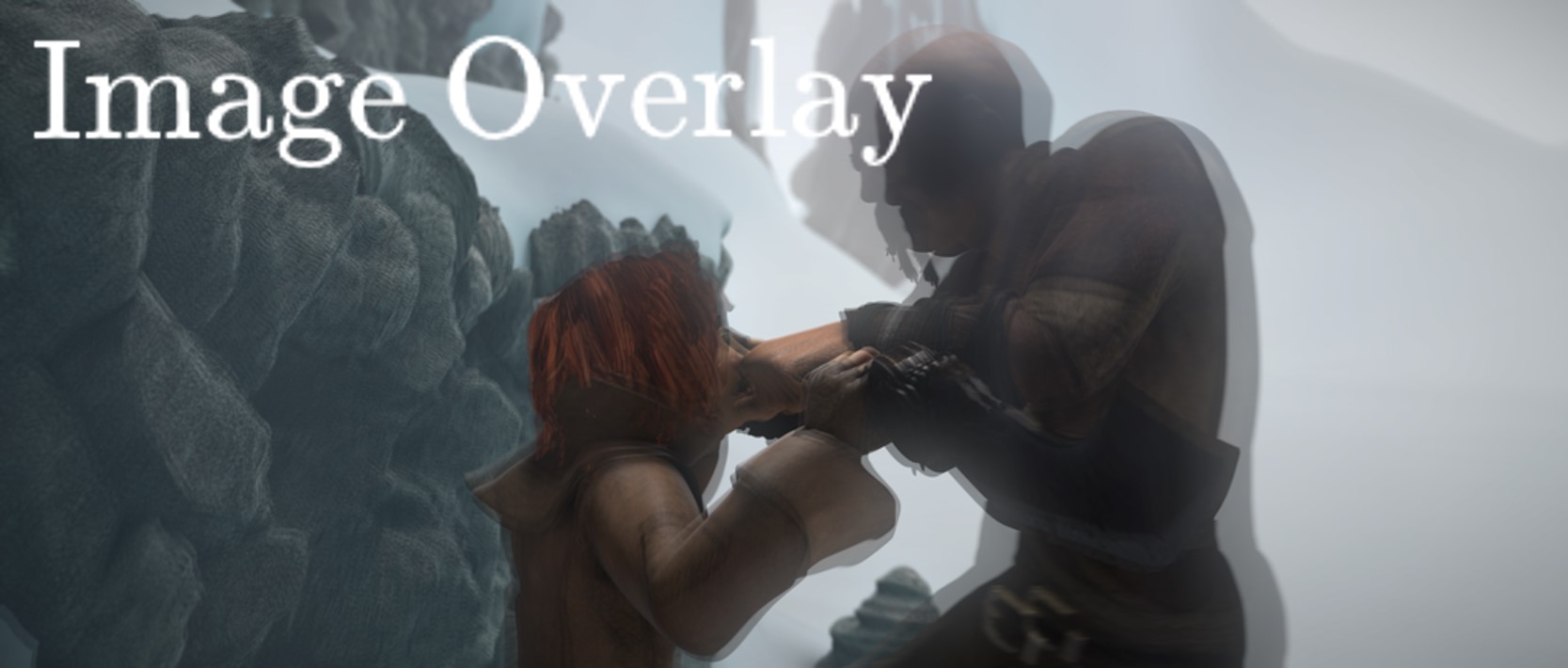}
& \hspace{-0.4cm} \includegraphics[width=4.2cm]{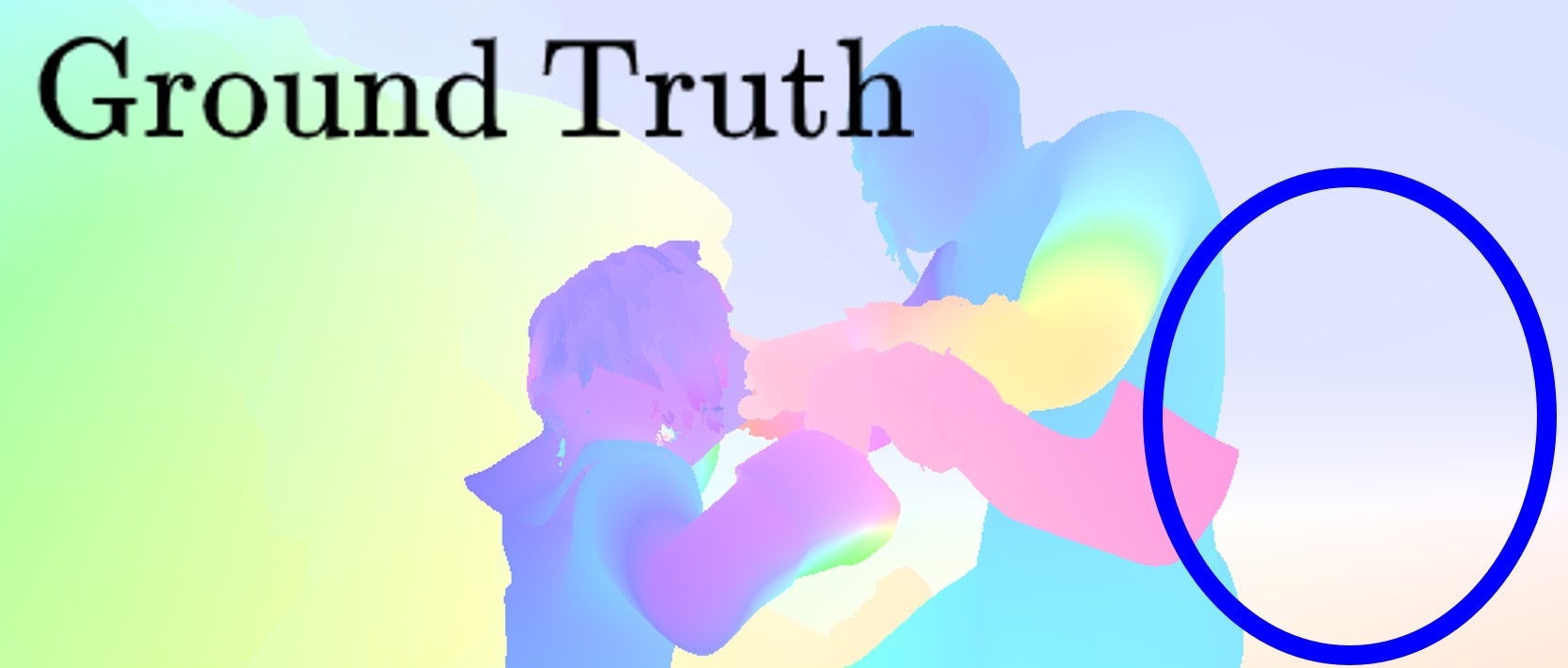}\\

\hspace{-0.2cm} \includegraphics[width=4.2cm]{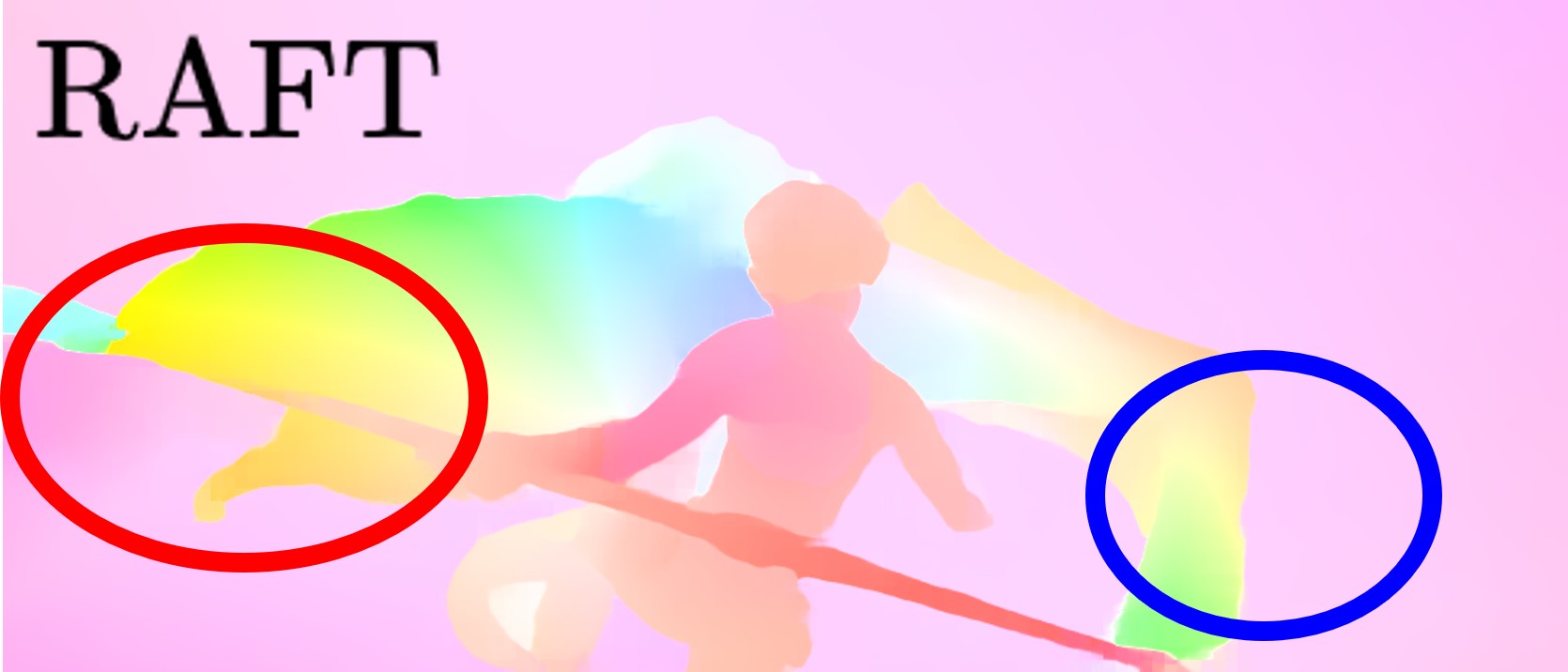}
& \hspace{-0.4cm} \includegraphics[width=4.2cm]{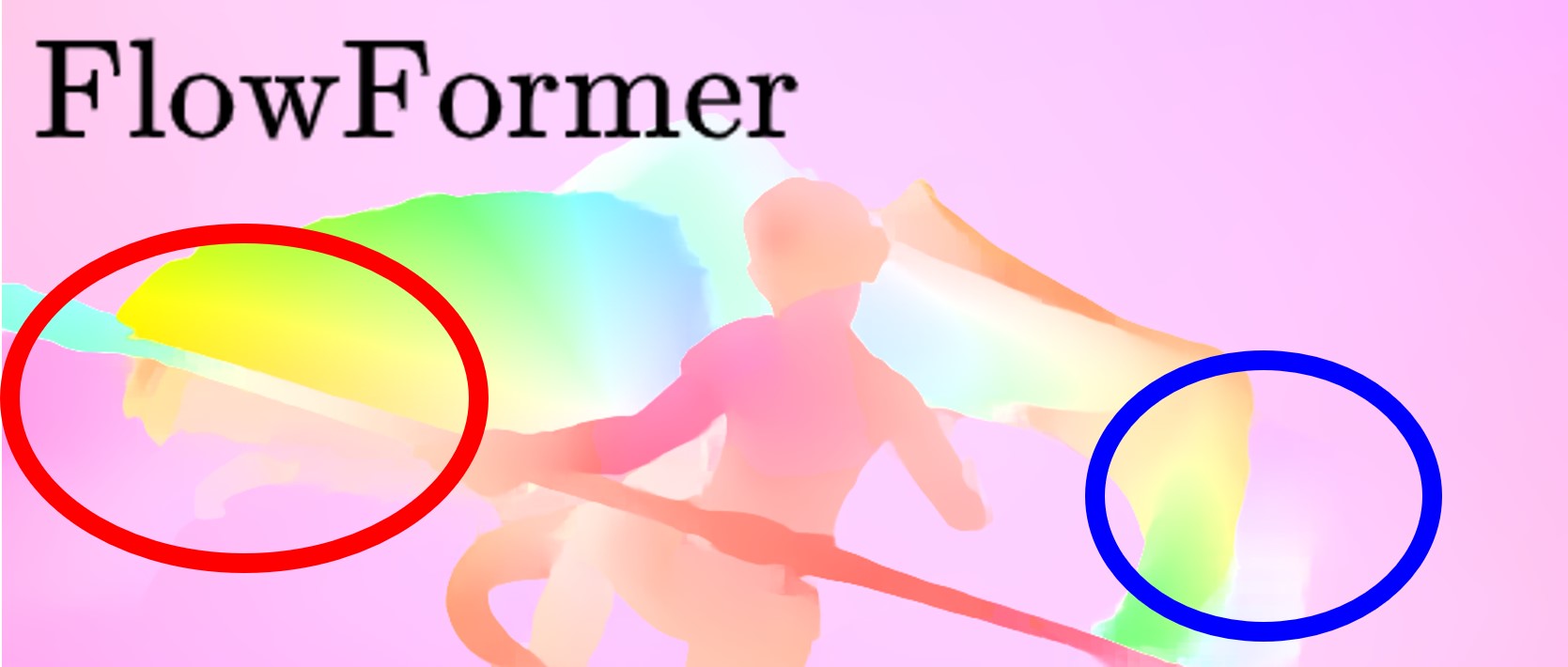}
& \hspace{-0.4cm} \includegraphics[width=4.2cm]{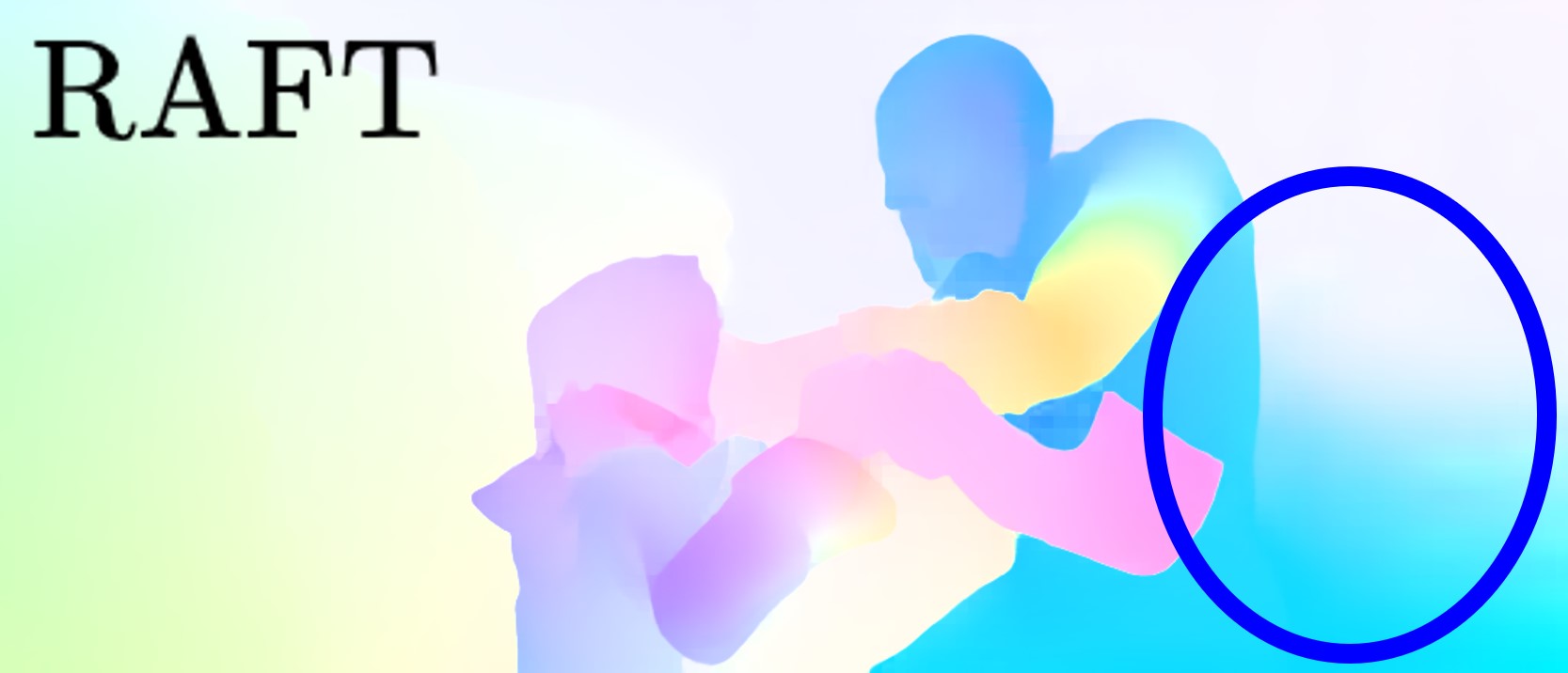}
& \hspace{-0.4cm} \includegraphics[width=4.2cm]{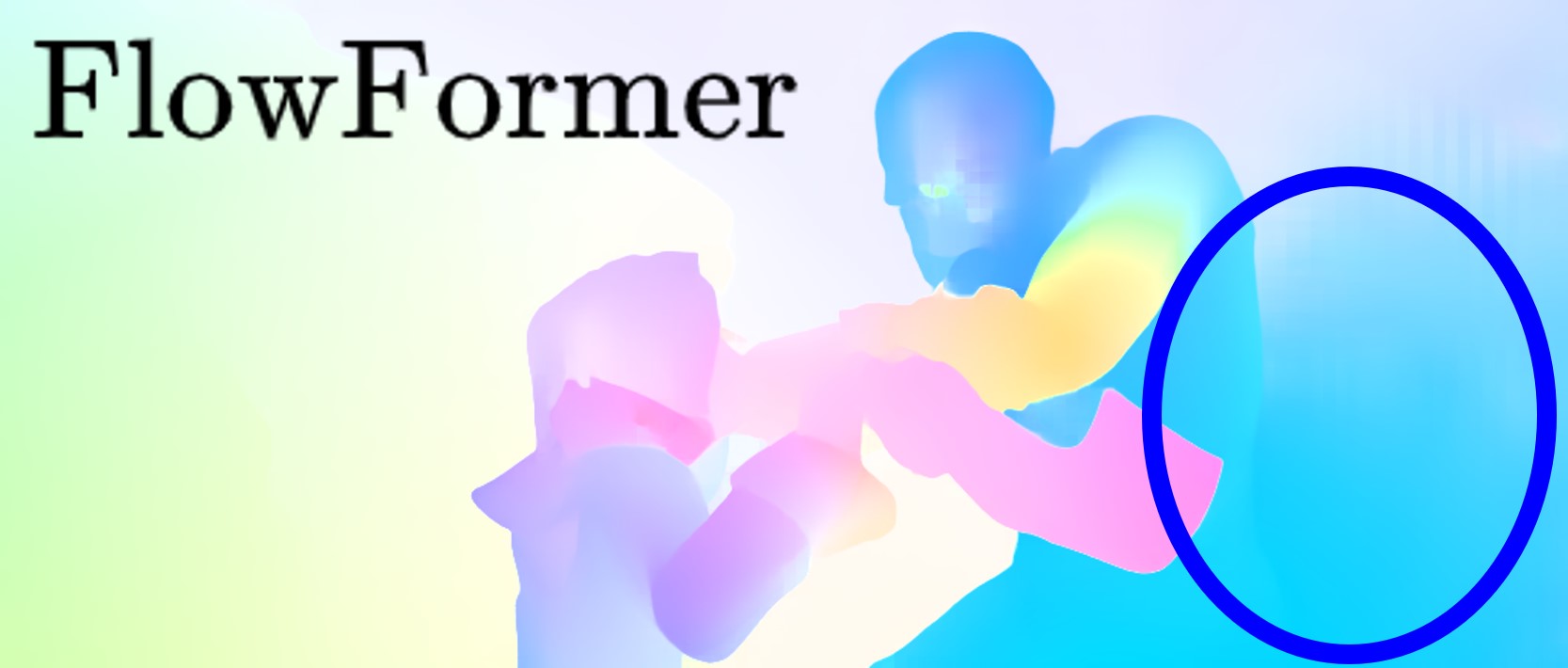}\\
\hspace{-0.2cm} \includegraphics[width=4.2cm]{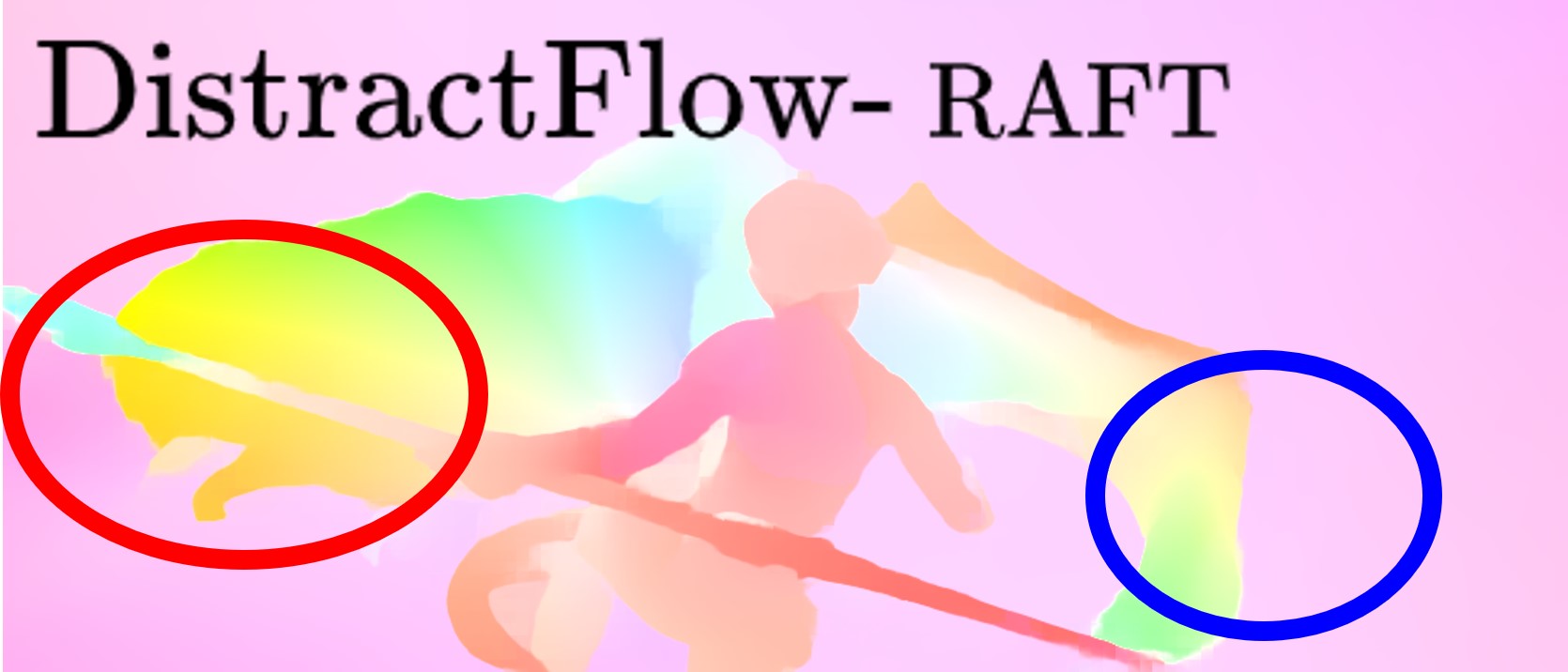}
& \hspace{-0.4cm} \includegraphics[width=4.2cm]{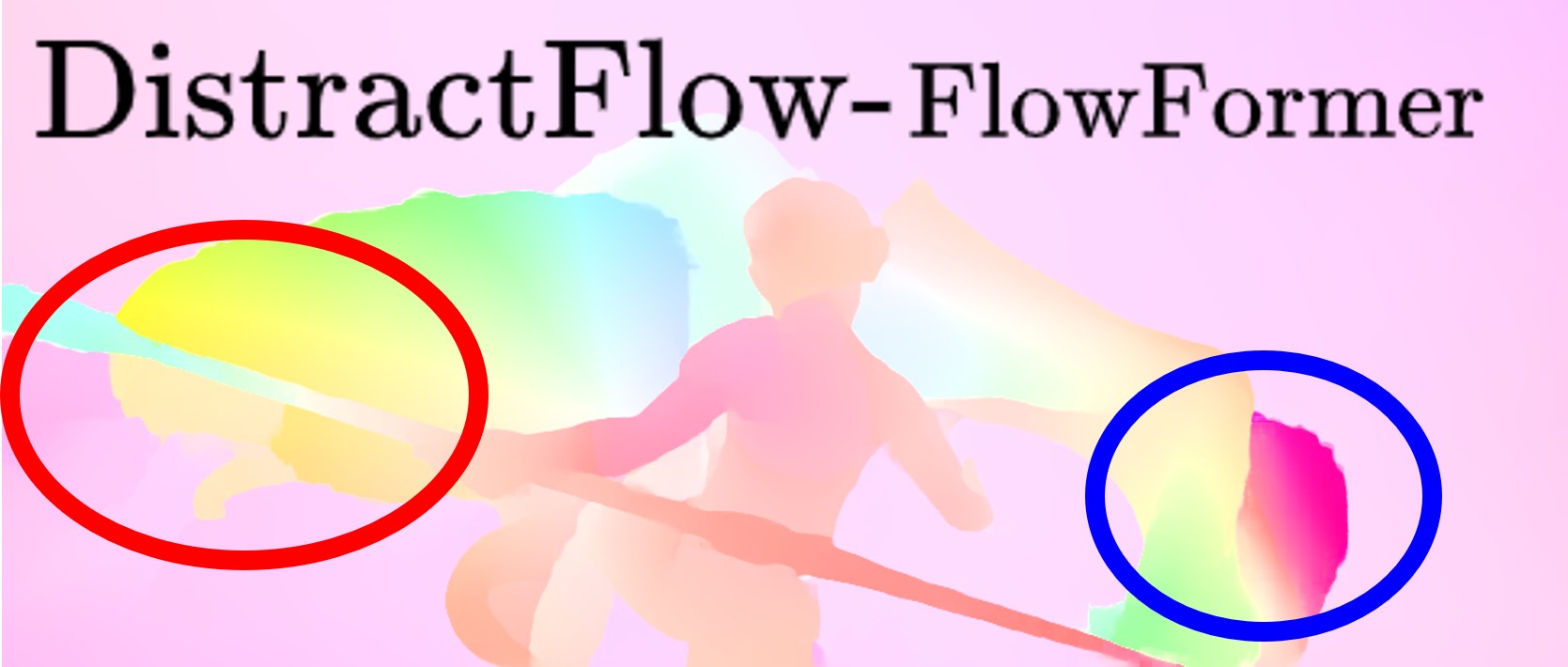}
&\hspace{-0.4cm} \includegraphics[width=4.2cm]{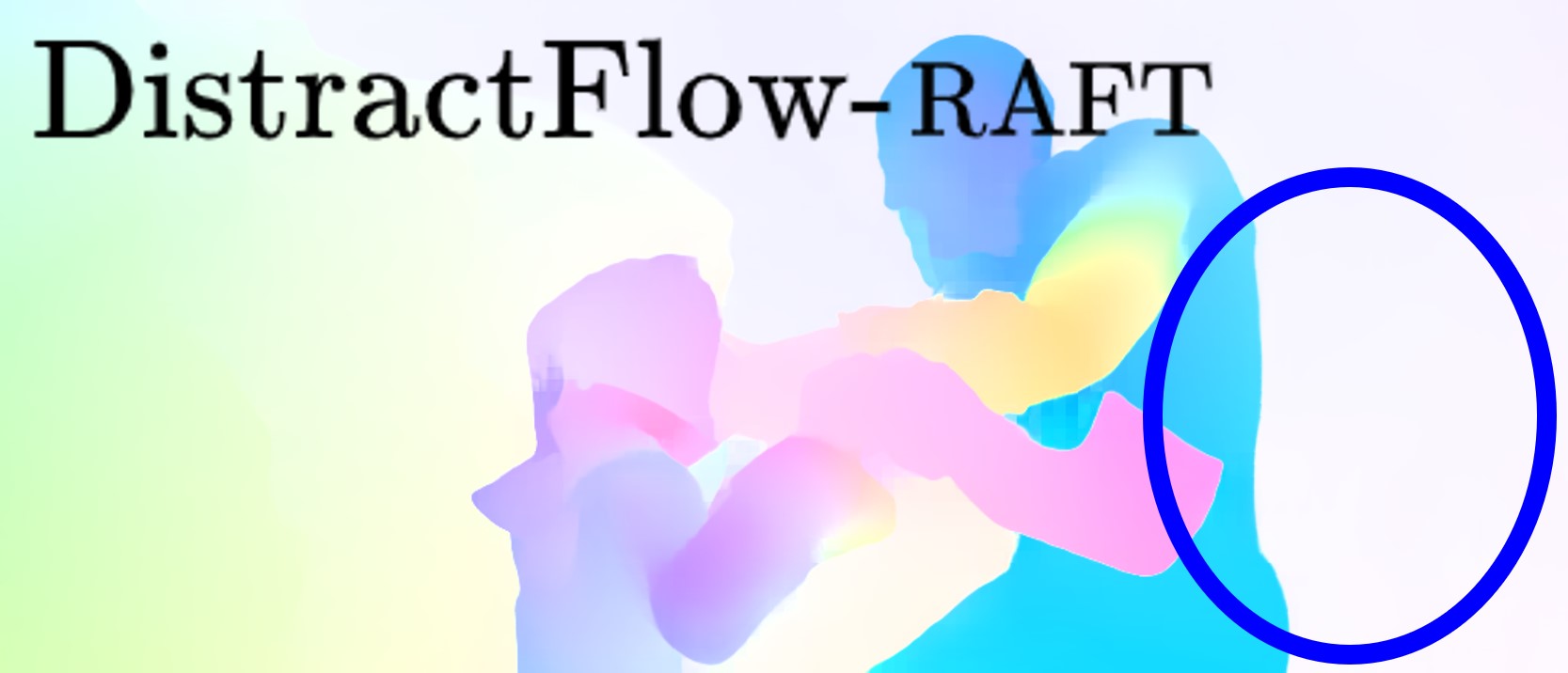}
& \hspace{-0.4cm} \includegraphics[width=4.2cm]{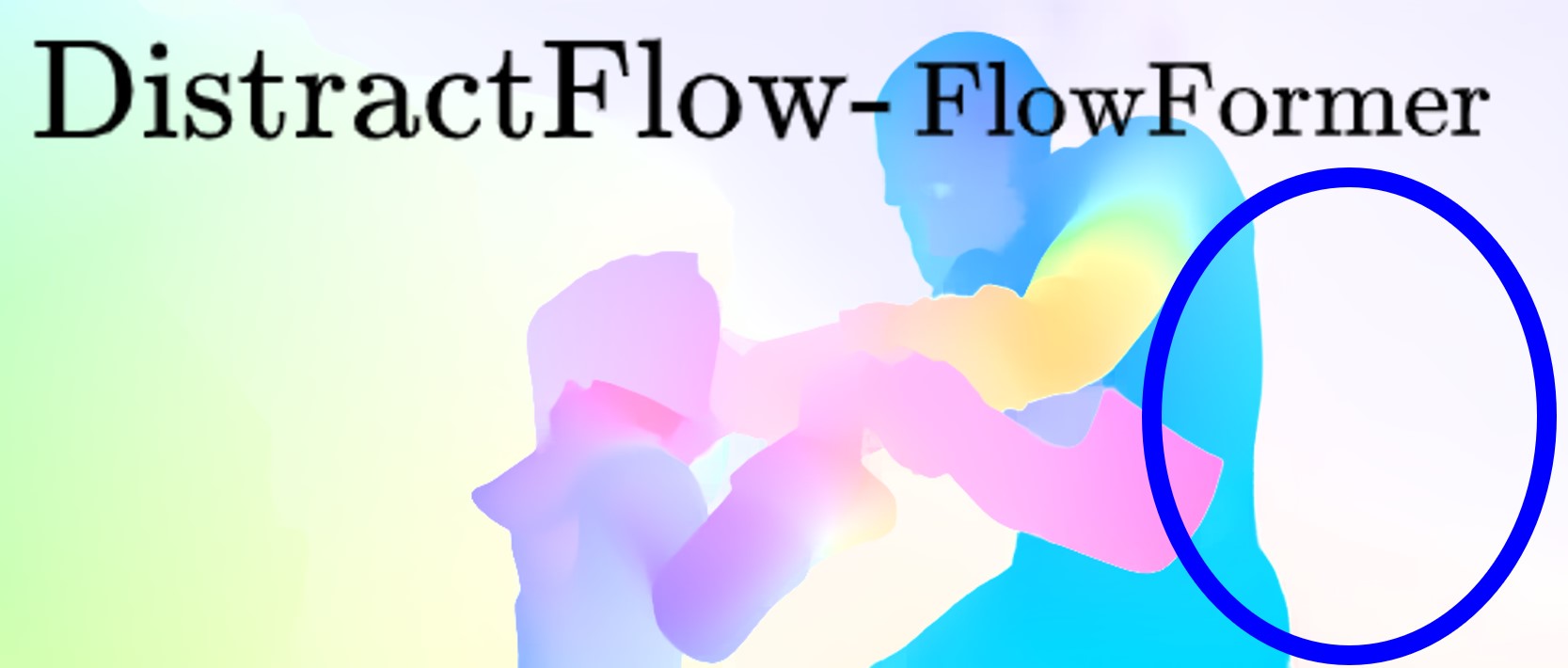}\\

% \hspace{-0.2cm} \includegraphics[width=4.2cm]{images/cave4_raft.jpg}
% & \hspace{-0.4cm} \includegraphics[width=4.2cm]{images/cave4_flowformer.jpg}
% & \hspace{-0.4cm} \includegraphics[width=4.2cm]{images/ambush5_raft.jpg}
% & \hspace{-0.4cm} \includegraphics[width=4.2cm]{images/ambush5_flowformer.jpg}\\
% \hspace{-0.2cm} \includegraphics[width=4.2cm]{images/cave4_raft_our.jpg}
% & \hspace{-0.4cm} \includegraphics[width=4.2cm]{images/cave4_flowformer_our.jpg}
% &\hspace{-0.4cm} \includegraphics[width=4.2cm]{images/ambush5_raft_our.jpg}
% & \hspace{-0.4cm} \includegraphics[width=4.2cm]{images/ambush5_flowformer_our.jpg}\\

\end{tabular}$
\end{center}
\vspace{-7mm}
\caption{Qualitative results on Sintel (train) using RAFT, FlowFormer, DistractFlow-RAFT, and DistractFlow-FlowFormer (the last two are empowered with our proposed method). All models are trained on FlyingChairs and FlyingThings3D. It can be seen that DistractFlow can generate more accurate predictions, with better spatial consistency and finer details, as highlighted by circles.  
%\fp{Herbert/Jisoo, please add a short sentence on our improvement here}
%\js{I added EPE in the image, but it seems messy , now could you check it please?} no worries, we have a lot of numbers in the paper
% \js{last row, DistractFlow-Flowformer shows large EPE compared to FlowFormer. Do we still add this example?}
}
\label{exp:qualitative_sintel}
\vspace{-10pt}
% \vspace{-1mm}
\end{figure*}

% Yes, is it okay if I remove it? Thanks, I will. ... fp: Thanks!

\subsection{Ablation Studies}\vspace{-3pt}
We conduct extensive ablation studies on various aspects of our method, by using RAFT as the base model. In the supervised setting, we train on FlyingChairs (C) and FlyingThings3D (T). In the semi-supervised setting, we take the RAFT model pretrained from the supervised setting and then finetune it using FlyingThings3D (labeled) and Sintel (test)/KITTI (test) (unlabeled). The evaluation is done on Sintel (train) and KITTI (train).

\begin{table}[t]
\centering
\caption{Effects of different types of perturbations applied to the frames, as data augmentation during training.}
\vspace{-3mm}
\label{tab:synvsreal}
\adjustbox{max width=0.48\textwidth}
{
\begin{tabular}{|l||c||c|c|c|}
\hline
\multirow{2}{*}{Method}  & \multirow{2}{*}{Perturbation} & Sintel (train)  & \multicolumn{2}{|c|}{KITTI (train)} \\
% \cline{2-3}
\cline{3-5}
&  &  (Final-epe) & (Fl-epe) & (Fl-all) \\

\hline
\hline
RAFT \cite{teed2020raft} &   & 2.73 & 4.94 & 16.9 \\
\hline
\multirow{3}{*}{Our} & Gaussian noise & 2.68 & 4.86 & 17.6\\
 & Random shapes & 2.66 & 4.82 & 16.8\\
 & Realistic Distractions & \textbf{2.61} & \textbf{4.57} & \textbf{16.4}\\
\hline
% \hline
\end{tabular}
}
\vspace{-8pt}
\end{table}

\begin{table}[t]
\begin{center}
\caption{Distracting $I_{t}$ and/or $I_{t+1}$ on Sintel (train) and KITTI (train) datasets. We train RAFT with distractions to $I_{1}$ or $I_{2}$ or both. $\alpha$ is the coefficient in the Beta distribution for sampling $\lambda$. $\alpha_{1}$ and $\alpha_{2}$ are for applying distractions to $I_{1}$ and $I_{2}$, respectively.}
\vspace{-3mm}
\label{tab:mix1vsmix2}
\adjustbox{max width=0.48\textwidth}
{
\begin{tabular}{|l||c|c|c||c|c|c|}
\hline

\multirow{2}{*}{Method}  & \multirow{2}{*}{Distraction} & \multirow{2}{*}{$\alpha_1$} & \multirow{2}{*}{$\alpha_2$} & Sintel (train)  & \multicolumn{2}{|c|}{KITTI (train)} \\

\cline{5-7}
&  & & & (Final-epe) & (Fl-epe) & (Fl-all) \\

% Method & Labeled & Network & \multicolumn{2}{|c|}{mAP (\%)}\\
% \multirow{2}{*}{Method} & Training  & \multicolumn{2}{|c|}{Sintel (train-EPE)}  & \multicolumn{2}{|c|}{KITTI (train)} & \multicolumn{2}{|c|}{Sintel (test-EPE)}  & KITTI (test) \\
% \cline{3-9}
% &  dataset & (Clean) & (Final) & (Fl-epe) & (Fl-all) & (Clean) & (Final) & (Fl-all) \\
\hline
\hline
RAFT \cite{teed2020raft} &  & &  & 2.73 & 4.94 & 16.9 \\
\hline
\multirow{8}{*}{DistractFlow} & \multirow{3}{*}{On $I_{t}$} & 0.1 &   & 2.69 & \textbf{4.81} & \textbf{16.4} \\
&   & 1 &   & 2.64 & 5.25 & 17.6 \\
&   & 10 &   & \textbf{2.55} & 5.32 & 17.8 \\
\cline{2-7}
& \multirow{3}{*}{On $I_{t+1}$} &  & 0.1  & 2.65 & 4.66 & \textbf{16.3} \\
&  &  & 1  & \textbf{2.61} & \textbf{4.57} & 16.4 \\
&  & & 10 & 2.70 & 4.82 & 17.2 \\
\cline{2-7}
&  On $I_{t}$ \& $I_{t+1}$ (same) & 0.1 & 1  & 2.70 & 5.33 & 18.3 \\
&  On $I_{t}$ \& $I_{t+1}$ (diff) & 0.1 & 1  & \textbf{2.64} & \textbf{4.92} & \textbf{16.7} \\
\hline

\hline
\end{tabular}
}
\end{center}
\vspace{-20pt}
\end{table}

% \subsection{Ablation studies for Synthetic Noise vs. Real Image Distraction}
\textbf{Type of Perturbations:}
When applying visual perturbations, we compare using realistic distractions in DistractFlow with using synthetic noise such as Gaussian noise and random shapes. For generating random shapes, we make random background colors and add 5 to 10 shapes (e.g., circles, triangles, and rectangles) of random colors and sizes. As shown in Table~\ref{tab:synvsreal}, while introducing perturbations with Gaussian noise or random shapes can result in performance gains, the improvements are not as significant as compared to the case of using our proposed augmentation strategy.

% \subsection{Ablation studies for Distract $I_{1}$ vs. Distract $I_{2}$ }
\textbf{Distracting $I_{t}$ or $I_{t+1}$ or both:}
We compare applying distractions to $I_{t}$ or $I_{t+1}$ or both in Table.~\ref{tab:mix1vsmix2}. At the top of Table.~\ref{tab:mix1vsmix2}, we generate distracted $I_{t}$ using the $\alpha_1$ values for sampling $\lambda$ and train the model accordingly.\footnote{When $\alpha < 1$, sampled $\lambda$ is close to 0 or 1. When $\alpha > 1$, sampled $\lambda$ is close to 0.5. When $\alpha = 1$, $\lambda$ is sampled from a uniform distribution.} All three variants show improvements on Sintel. However, when $\alpha_{1}$s are 1 or 10, the performance drops on KITTI compared to RAFT. We suspect that the Sintel (final) dataset has a heavy visual effect even in $I_{t}$, and using strongly distracted $I_{t}$ could help the training.\footnote{Since we set $w_{dist}=\lambda$, when $\lambda$ is close to zero, the distracted pair only impacts the training minimally.} On the other hand, since KITTI has relatively cleaner images, strongly distracted $I_{t}$ does not improve the performance.
% It shows that minor distraction to the $I_{1}$ is helpful.
% It shows that small augmentation to the $I_{1}$ is helpful, but large augmentation hurt the performance. 

In the middle of Table~\ref{tab:mix1vsmix2}, we distract $I_{t+1}$ according to the different $\alpha_2$ values and train the model. In most of these cases DistractFlow improves upon the baseline RAFT, and it shows the best performance at $\alpha_2=1$. This result shows that various mixing of the images is helpful. 

\begin{table}[t]
\begin{center}
\caption{Effectiveness of $L_{dist}$}
% on Sintel (train) and KITTI (train) dataset. We trained C+T dataset using RAFT architecture, which applied $L_{dist}$ loss (w/wo $L_{base}$ loss).
\vspace{-3mm}
\label{tab:sup_weight}
\adjustbox{max width=0.45\textwidth}
{
\begin{tabular}{|l||c|c||c|c|c|}
\hline

\multirow{2}{*}{Method}   & \multirow{2}{*}{$\mathcal{L}_{base}$} & \multirow{2}{*}{$\mathcal{L}_{dist}$} & Sintel (train) & \multicolumn{2}{|c|}{KITTI (train)} \\
% \cline{2-3}
\cline{4-6}
&  &  & (Final-epe) & (Fl-epe) & (Fl-all) \\

\hline
\hline
RAFT \cite{teed2020raft} & \checkmark & & 2.73 & 4.94 & 16.9\\
\hline
% \multirow{4}{*}{DistractFlow} &   & 1 & 2.78 & 5.63 & 19.4 \\
\multirow{2}{*}{DistractFlow}  & & \checkmark & 2.82 & 5.43 & 19.0 \\
 \cline{2-6}
%  & \checkmark & 1  & 2.66 & 5.02 & 17.4 \\
 & \checkmark  & \checkmark & \textbf{2.61} & \textbf{4.57} & \textbf{16.4} \\
 
\hline
\end{tabular}
}
\end{center}
\vspace{-20pt}
\end{table}

At the bottom of Table~\ref{tab:mix1vsmix2}, we distract $I_{t}$ and $I_{t+1}$ simultaneously. When we apply the same distraction to both frames, two consecutive videos, it degrades the performance. This is because applying the same distraction to both frame introduces new correspondences which are not part of the ground truth. On the other hand, when we apply distractions from two different images to $I_{t}$ and $I_{t+1}$, the trained model shows better performance compared to baseline RAFT. Overall, distracting $I_{t+1}$ shows the best performance and we carry out experiments with this setting.

% We compare applying distractions to $I_{1}$ or $I_{2}$ in Table.~\ref{tab:mix1vsmix2}. At the top of Table.~\ref{tab:mix1vsmix2}, we generated distract $I_{1}$ according to the change of the $\alpha$\footnote{When $\alpha < 1$, sampled $\lambda$ is close to 0 or 1. \\ When $\alpha > 1$, sampled $\lambda$ is close to 0.5. \\ When $\alpha = 1$, $\lambda$ is uniform distribution} for $\lambda$ and trained the model. All three model shows improvement on Sintel. But, when $\alpha_{1}$s are 1 or 10, it drops the performance on KITTI compared to RAFT. It shows that small augmentation to the $I_{1}$ is helpful, but large augmentation hurt the performance. In the middle of Table.~\ref{tab:mix1vsmix2}, we distracted $I_{2}$ according to the different $\alpha$ and trained the model. Most of these cases show better performance compared to RAFT, and it shows the best performance at $\alpha$=1. This result shows that various mixing of the images is helpful. At the bottom of Table.~\ref{tab:mix1vsmix2}, we generated either distract $I_{1}$ and $I_{2}$ simultaneously. When we apply the same distraction to two consecutive videos, it degrades the performance. It can have two correspondence, which can have problems. On the other hand, when we generate the distract images from different images for $I_{1}$ and $I_{2}$, it shows better performance compared to RAFT. In overall, distracting $I_{2}$ shows the best performance and we have experimented with this setting. 

% \subsection{Effectiveness of $L_{dist}$}
\textbf{Effectiveness of $\mathcal{L}_\textbf{dist}$:}
Table~\ref{tab:sup_weight} shows the effectiveness of $\mathcal{L}_\text{dist}$. Without the supervised loss on the original pairs, a model trained only with $\mathcal{L}_\text{dist}$ still provides reasonable estimation performance, but shows worse accuracy as compared to the original model. Combining $\mathcal{L}_\text{base}$ with $\mathcal{L}_\text{dist}$ shows a significant improvement.

% And, without $w_{dist}$ weight, it still shows similar performance with baseline. However, applying $w_{dist}$ weight shows significant improvement compared to baseline. Weighting them by the mixing ratio can account for cases where $\lambda$ is very close to zero or one. Therefore, the weight is very important for the augmented training for optical flow estimation.

\begin{table}[t]
\vspace{0pt}
\begin{center}
\caption{Effectiveness of confidence map and $\lambda$ weight for $\mathcal{L}_\text{self}$ in semi-supervised setting. $\tau$ is the confidence threshold in Eq.~\ref{eq:semi_1} and in $w_\text{self}$ is the weight for the self-supervised loss. Note that 0.37 corresponds to having equal nominator and denominator in Eq.~\ref{For-Back_Con}.}
\vspace{-3mm}
\label{tab:conf_map}
\adjustbox{max width=0.42\textwidth}
{
\begin{tabular}{|l|c||c|c|c|}
\hline

\multirow{2}{*}{Method}  & \multirow{2}{*}{$\tau$} &  Sintel (train)  & \multicolumn{2}{|c|}{KITTI (train)} \\
\cline{3-5}
& & (Final-epe) & (Fl-epe) & (Fl-all) \\

% Method & Labeled & Network & \multicolumn{2}{|c|}{mAP (\%)}\\
% \multirow{2}{*}{Method} & Training  & \multicolumn{2}{|c|}{Sintel (train-EPE)}  & \multicolumn{2}{|c|}{KITTI (train)} & \multicolumn{2}{|c|}{Sintel (test-EPE)}  & KITTI (test) \\
% \cline{3-9}
% &  dataset & (Clean) & (Final) & (Fl-epe) & (Fl-all) & (Clean) & (Final) & (Fl-all) \\
\hline
\hline
RAFT \cite{teed2020raft} & & 2.73 & 4.94 & 16.9 \\
% \hline
% FS \cite{im2022semi} (RAFT) & &  & 1.30 & 2.46 & 3.35 & 11.12\\
\hline
\multirow{3}{*}{DistractFlow} & & \multicolumn{3}{c|}{diverged} \\
%  & &  & \multicolumn{3}{c|}{diverged} \\
\cline{2-5}
& \checkmark (0.37)  & \textbf{2.35} & 3.37 & 12.42 \\
& \checkmark (0.95) & \textbf{2.35} & \textbf{3.01} & \textbf{11.71} \\
\hline
\end{tabular}
}
\end{center}
\vspace{-22pt}
% \vspace{-8mm}
% \vspace{-17pt}
\end{table}

\begin{table}[t]
\vspace{2mm}
\begin{center}
\caption{Effects of using different unlabeled datasets in  the semi-supervised training. Our DistractFlow enables consistent performance improvements when using any of the unlabeled datasets.}
\vspace{-3mm}
\label{tab:unlabel}
\adjustbox{max width=0.43\textwidth}
{
\begin{tabular}{|l|c||c|c|}
\hline

\multirow{2}{*}{Method}  & Unlabeled data & Sintel (train) \\
\cline{3-3}
& (\# of pairs) & (Final-epe)  \\

\hline
\hline
RAFT \cite{teed2020raft} & & 2.73 \\
\hline
\multirow{3}{*}{DistractFlow} & Sintel-test (1.1k) & \textbf{2.35}\\
& Monkaa (17k) \& Driving (9k) & 2.42\\
& Big Buck Bunny (14k) & 2.58\\

\hline
\end{tabular}
}
\end{center}
\vspace{-7mm}
\end{table}

\begin{figure}[t]
\begin{center}$
\centering
\begin{tabular}{c c }

% \hspace{-0.2cm} 
\includegraphics[width=3.9cm]{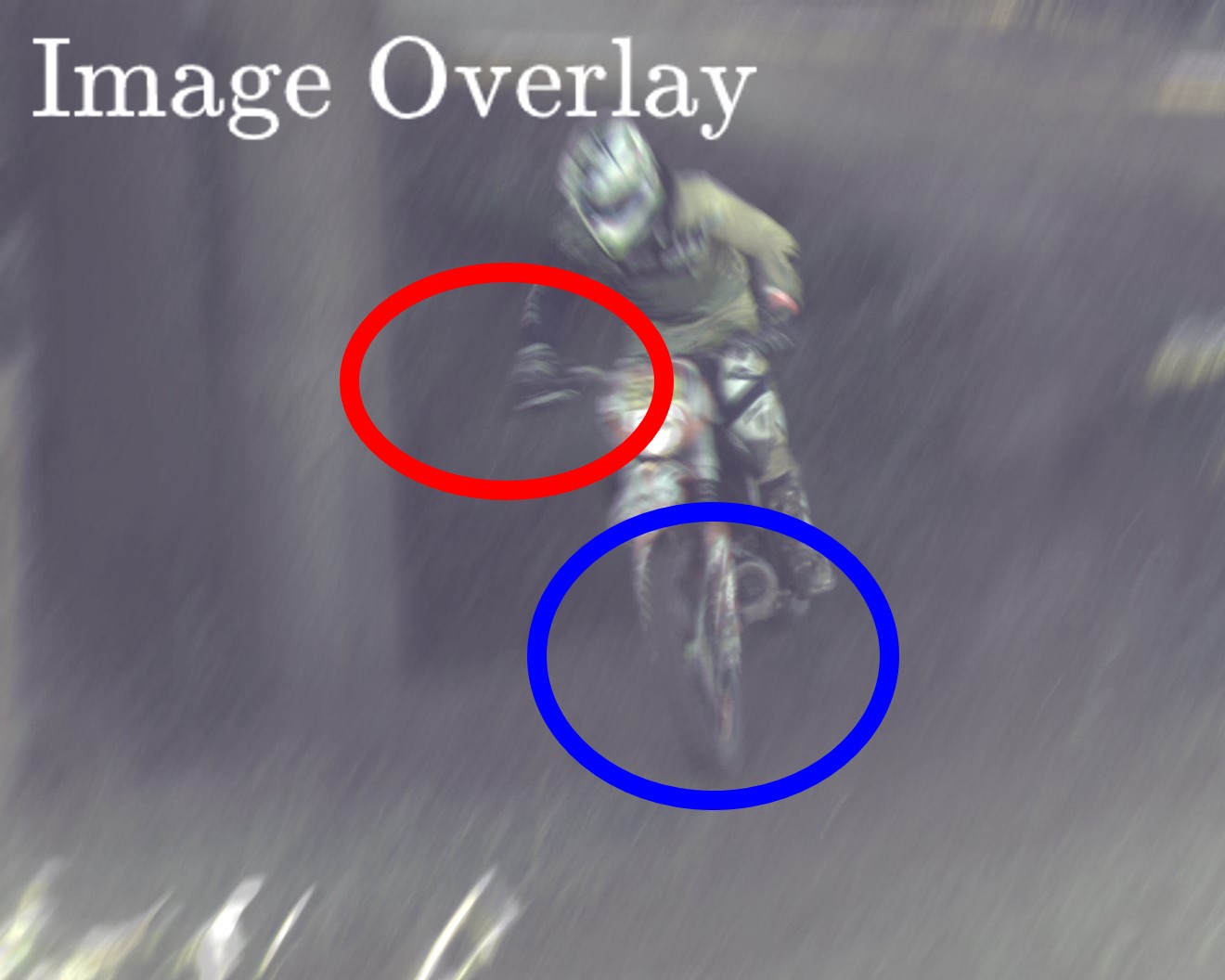} & 
\hspace{-0.4cm}
\includegraphics[width=3.9cm]{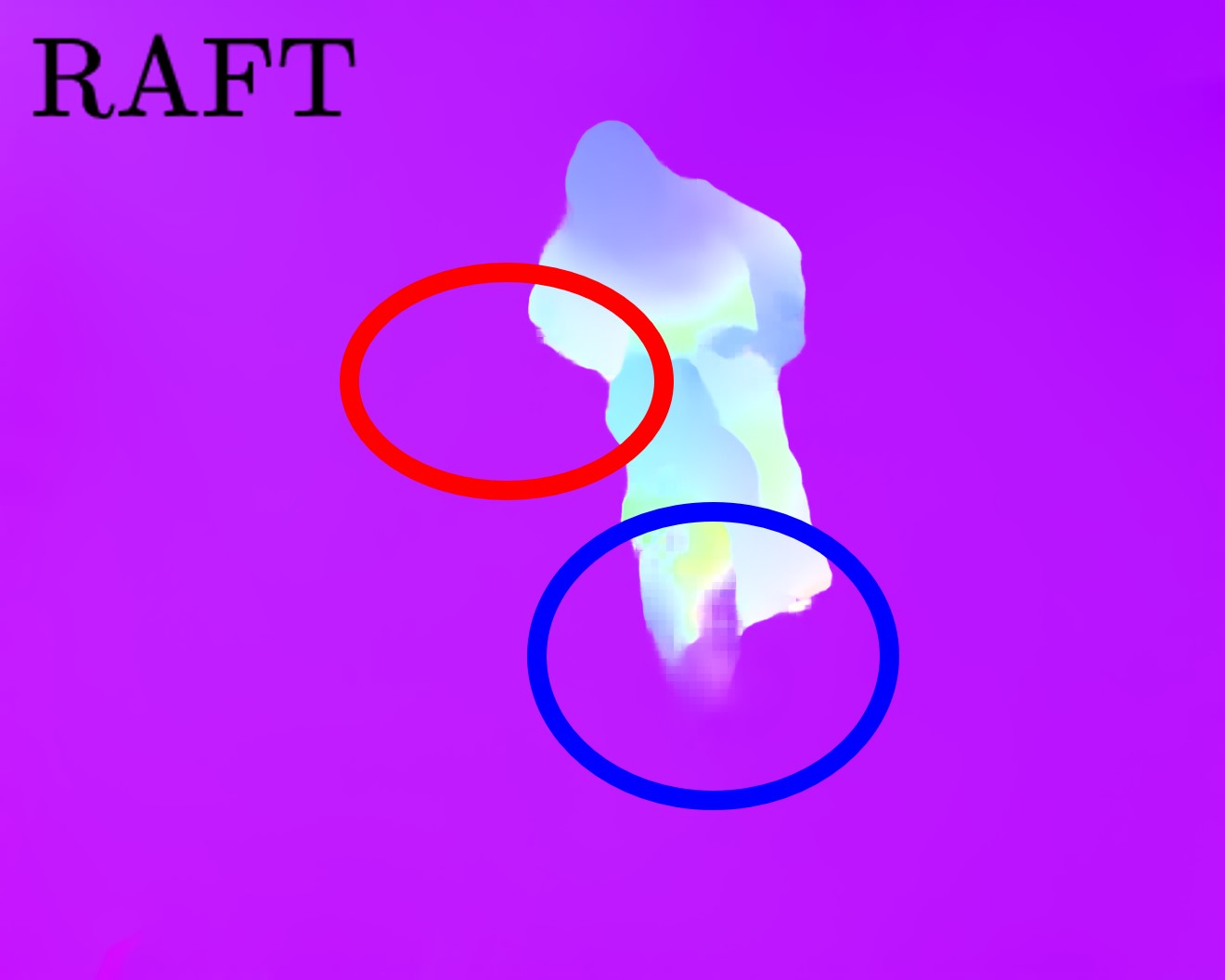}\\
%  \hspace{-0.2cm}
 
\includegraphics[width=3.9cm]{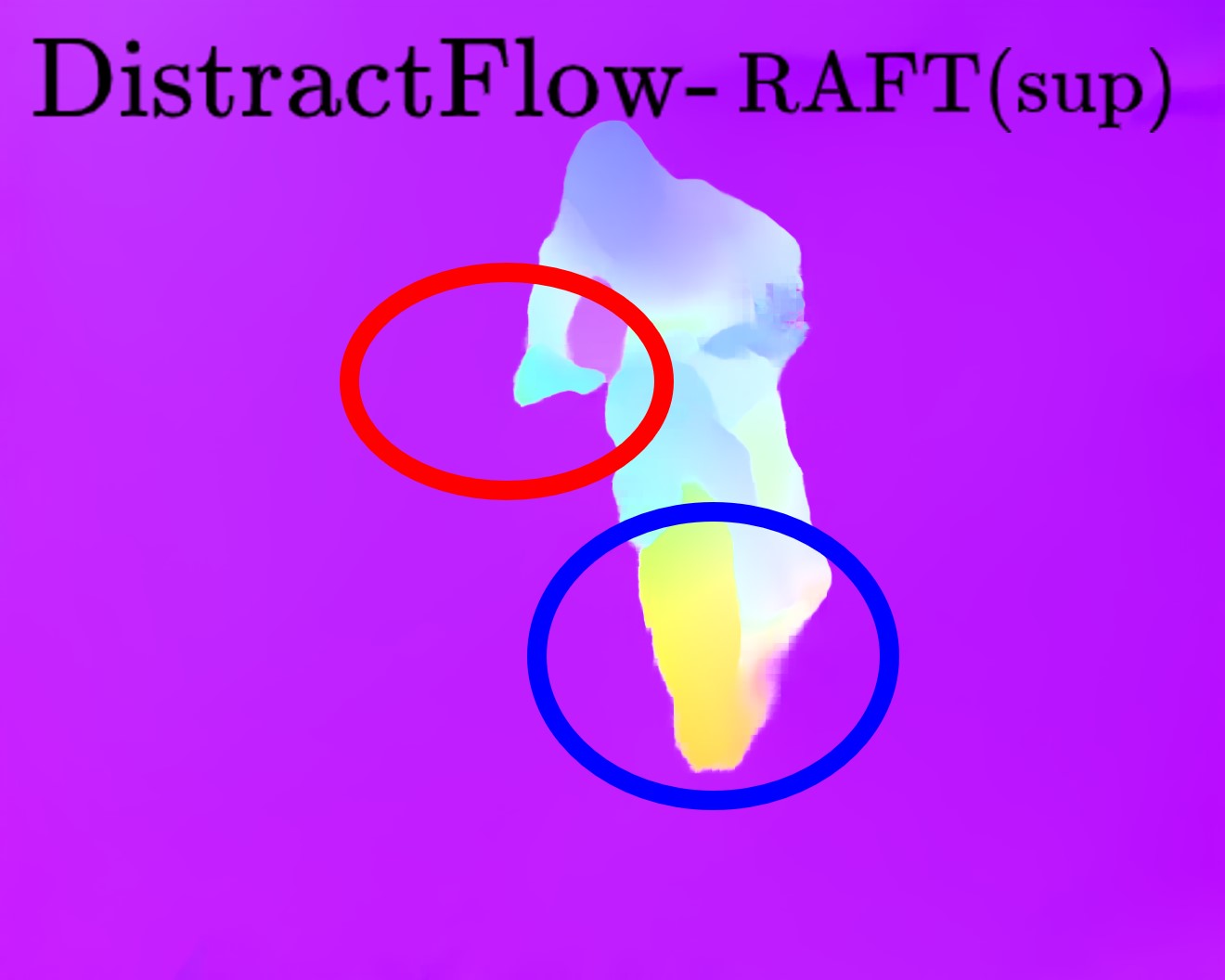} & 
\hspace{-0.4cm}
\includegraphics[width=3.9cm]{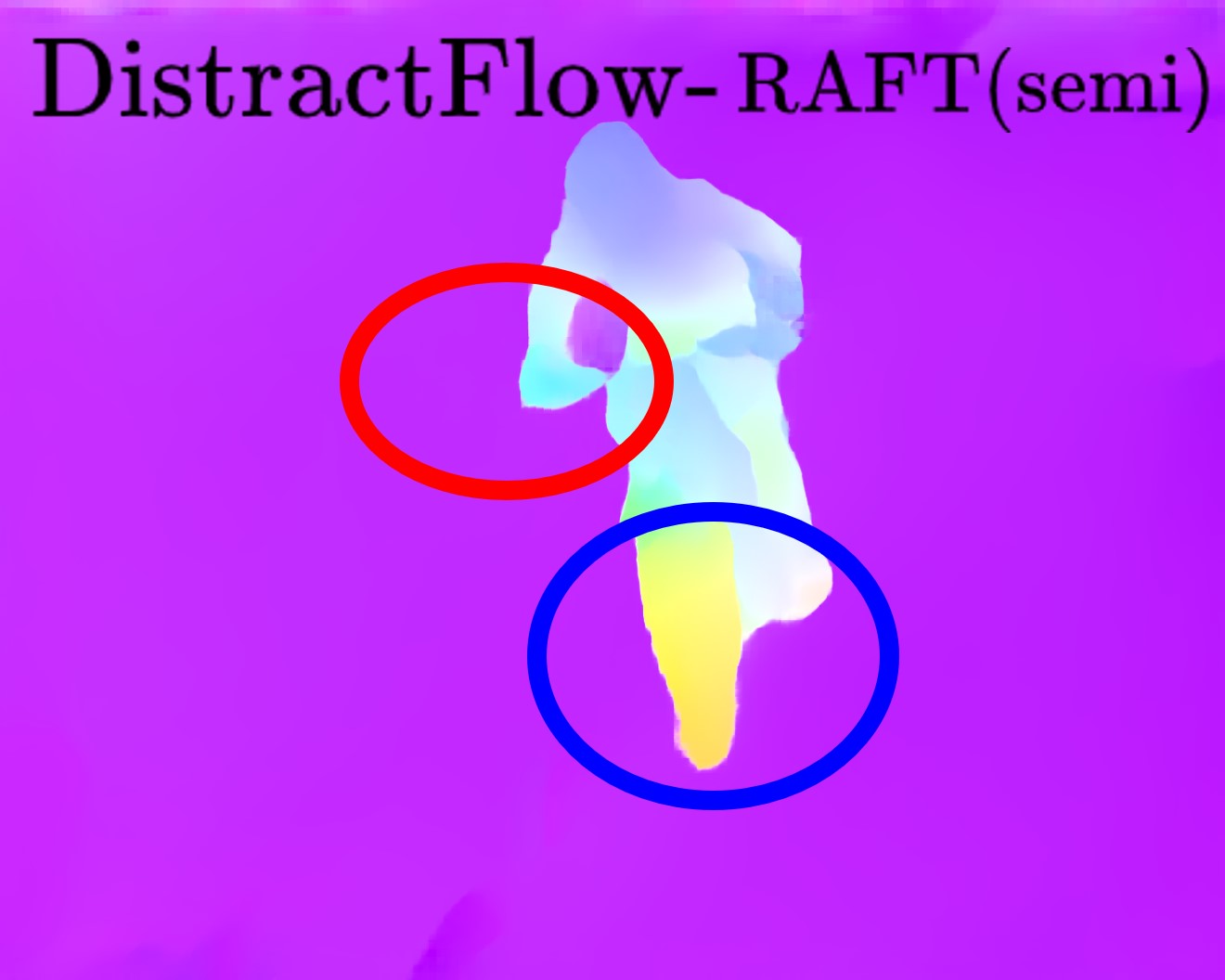}\\

\end{tabular}$
\end{center}
\vspace{-7mm}
\caption{Qualitative results on SlowFlow using the original RAFT and DistractFlow-trained RAFT models in supervised and semi-supervised model settings. We see that DistractFlow training enables the network to produce more accurate results, despite the severe motion blur. %The circles highlight sample regions with considerably improved flow predictions.
% (trained with C+T (label) and Sintel and KITTI (unlabel)). 
% \vspace{-3mm}
}
\label{exp:qualitative_kitti}
% \vspace{-1mm}
\vspace{-10pt}
\end{figure}

% \subsection{Effectiveness of Confidence Map ($M_{conf}$)}
\textbf{Effectiveness of Confidence Map:}
Table~\ref{tab:conf_map} shows our study on the impacts of confidence-based thresholding in the semi-supervised training. When training the model without any confidence maps, erroneous predictions are used for backpropagation and training the network, causing the training to diverge. On the other hand, our confidence map allows the network to only  train on highly accurate predictions and enables stability when training the model.

When we set $\tau=0.37$ (a common choice for forward-backward consistency), we can stably train the model and it shows improvement. Higher $\tau$ (e.g., 0.95) shows further accuracy improvement on KITTI. This may be due to the larger displacements in KITTI, which can make the model more prone to in accurate predictions. As such, creating pseudo labels with a higher confidence threshold is beneficial in this case.
% When we train the model without the confidence map, the model diverges because of outlier training. However, our confidence map helps the network train high accurate pseudo-label and it makes better performance than the previous semi-supervised learning method. When we set $\tau=0.37$ (original forward-backward consistency-based occlusion mask), it shows the same performance on the Sintel dataset, but higher $\tau$ shows further improvement on the KITTI dataset. We suspect that the KITTI unlabeled dataset has larger displacements than the Sintel dataset, making pseudo ground truth with a high threshold effectively suppressing inaccurate samples.

\subsection{Qualitative Results}\vspace{-3pt}
% \textbf{Qualitative Results. }
Figure~\ref{exp:qualitative_sintel} shows qualitative results on the Sintel (train, final) using the original RAFT and FlowFormer, as well as our DistractFlow-trained RAFT and FlowFormer. These models are trained on FlyingChairs and FlyingThings3D. Because Sintel (final) contains visual effects such as fog and blur, the original models generate erroneous estimations. On the other hand, our DistractFlow-trained models show more accurate and robust flow estimation results. Especially, it shows accurate predictions at the object boundary without any edge- or segmentation-aware training~\cite{cai2021x, borse2021inverseform}.

Figure~\ref{exp:qualitative_kitti} shows qualitative results on SlowFlow using the original RAFT, as well as DistractFlow-trained RAFT models (supervised and semi-supervised). Our supervised training allows the model to generate more accurate and robust flows compared to the baseline RAFT. With our semi-supervised setting, our model shows further improvements. 
% Fig.~\ref{exp:qualitative_kitti} shows the qulitative results for KITTI using RAFT and DistractFlow (Supervised, Semi-supervised settings). Using our supervised training, it reduces the error and more robust compared to RAFT. With our semi-supervised setting, our model shows further improvements. 

% Fig.~\ref{exp:qualitative} shows the qualitative results for Original and Distracted image pairs using RAFT and DistractFlow (RAFT) models trained on C+T. There are 3 mixing $I_{2}$ images and all of them has correspondence with $I_{1}^{a}$. When image 2 has small portion ($\lambda \le 0.5$) of $I_{2}^{a}$, our model shows much better performance than origial RAFT model. 

\section{Discussion}
\vspace{-3pt}

\textbf{Unlabeled Dataset:}
Table~\ref{tab:unlabel} shows the results on Sintel (train) when using different unlabeled data in semi-supervised training. Sintel (test) shows significant improvement compared to supervised training since it is in the same domain as Sintel (train). Although Monkaa \& Driving or Bunny dataset have more unlabeled pairs and can still improve upon the original RAFT, they exhibit worse performance than using Sintel (test) pairs. This indicates that for semi-supervised setting, it is important to use unlabeled data with scenes and distributions resembling the target use case.
We leave data distribution robustness (e.g., addressing out-of-distribution samples~\cite{vaze2022openset, liu2018open,liu2022pac} as part of future work.
Nevertheless, our semi-supervised method improves the performance when using any of the unlabeled datasets.

\begin{figure}[t]
\vspace{-2pt}
\begin{center}$
\centering
\begin{tabular}{c c c}
\small{\text{Images}} & \small{\text{RAFT}} & \small{\text{DistractFlow}}\\
% \text{Images} & \text{RAFT} & \text{DistractFlow}\\
\hspace{-0.2cm}
\includegraphics[width=2.6cm]{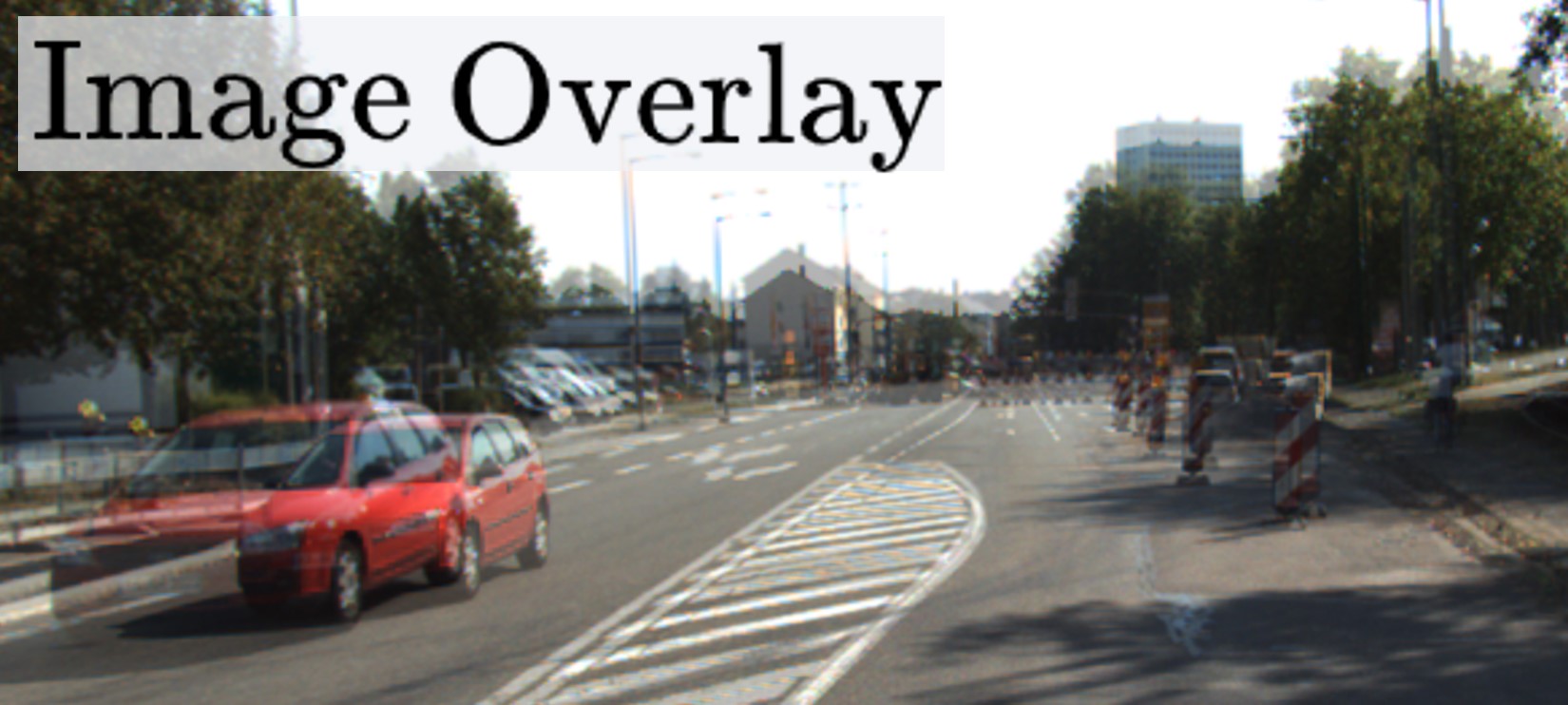} 
& \hspace{-0.4cm}
\includegraphics[width=2.6cm]{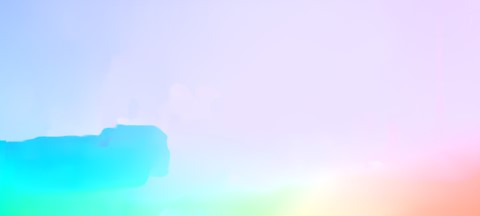}
& \hspace{-0.5cm}
\includegraphics[width=2.6cm]{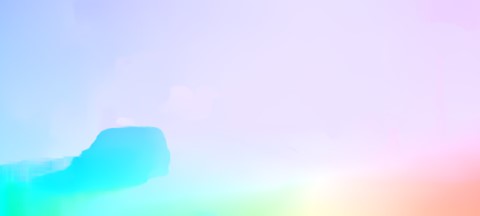}\\
\hspace{-0.2cm} 
\includegraphics[width=2.6cm]{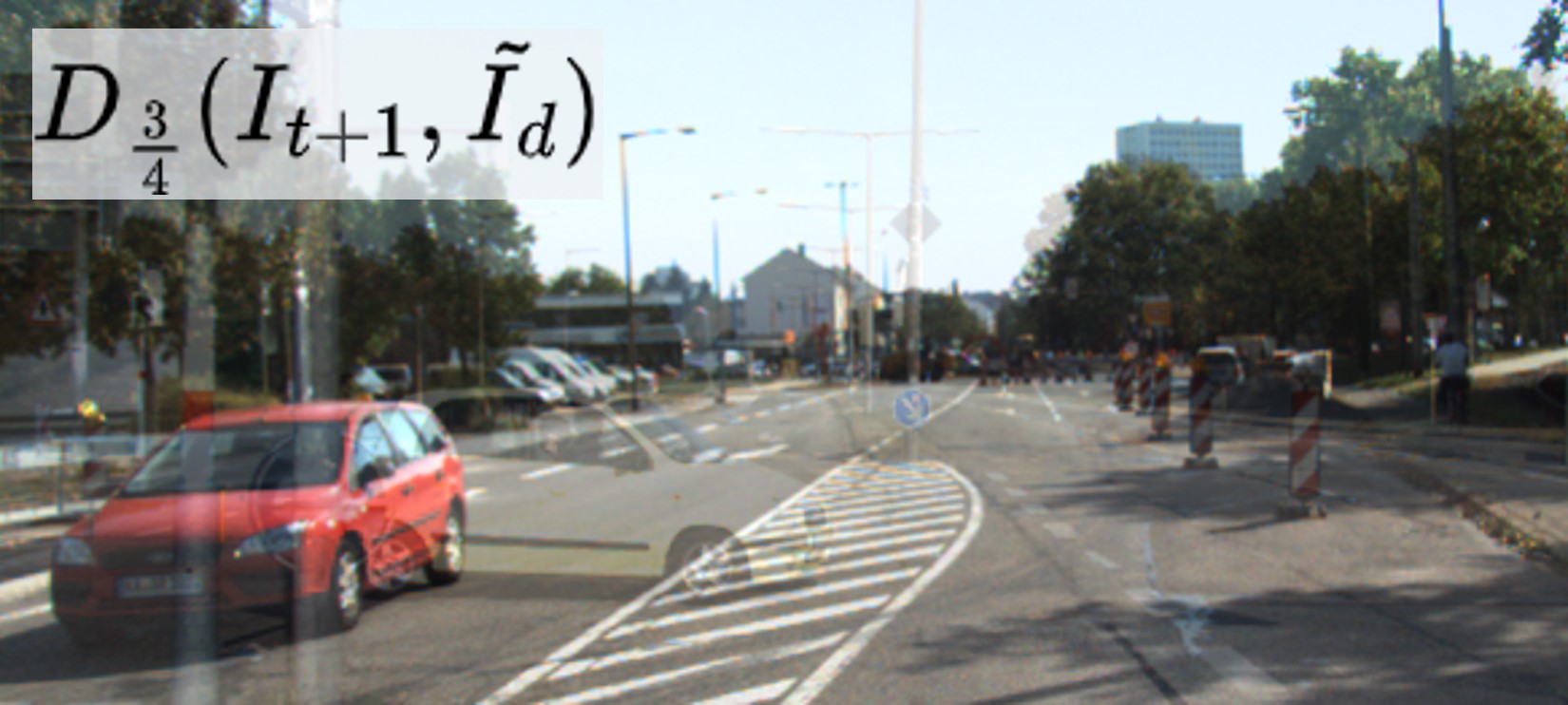}
& \hspace{-0.4cm}
\includegraphics[width=2.6cm]{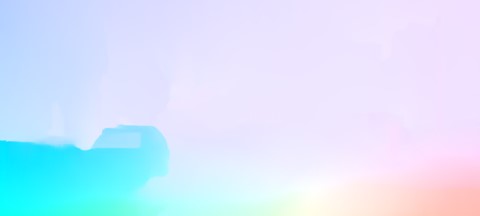} 
& \hspace{-0.5cm}
\includegraphics[width=2.6cm]{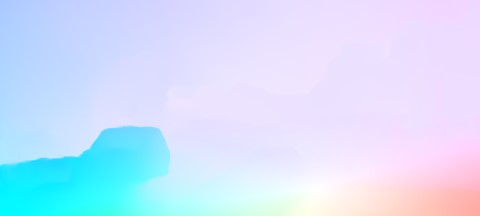}\\
\hspace{-0.2cm} 
\includegraphics[width=2.6cm]{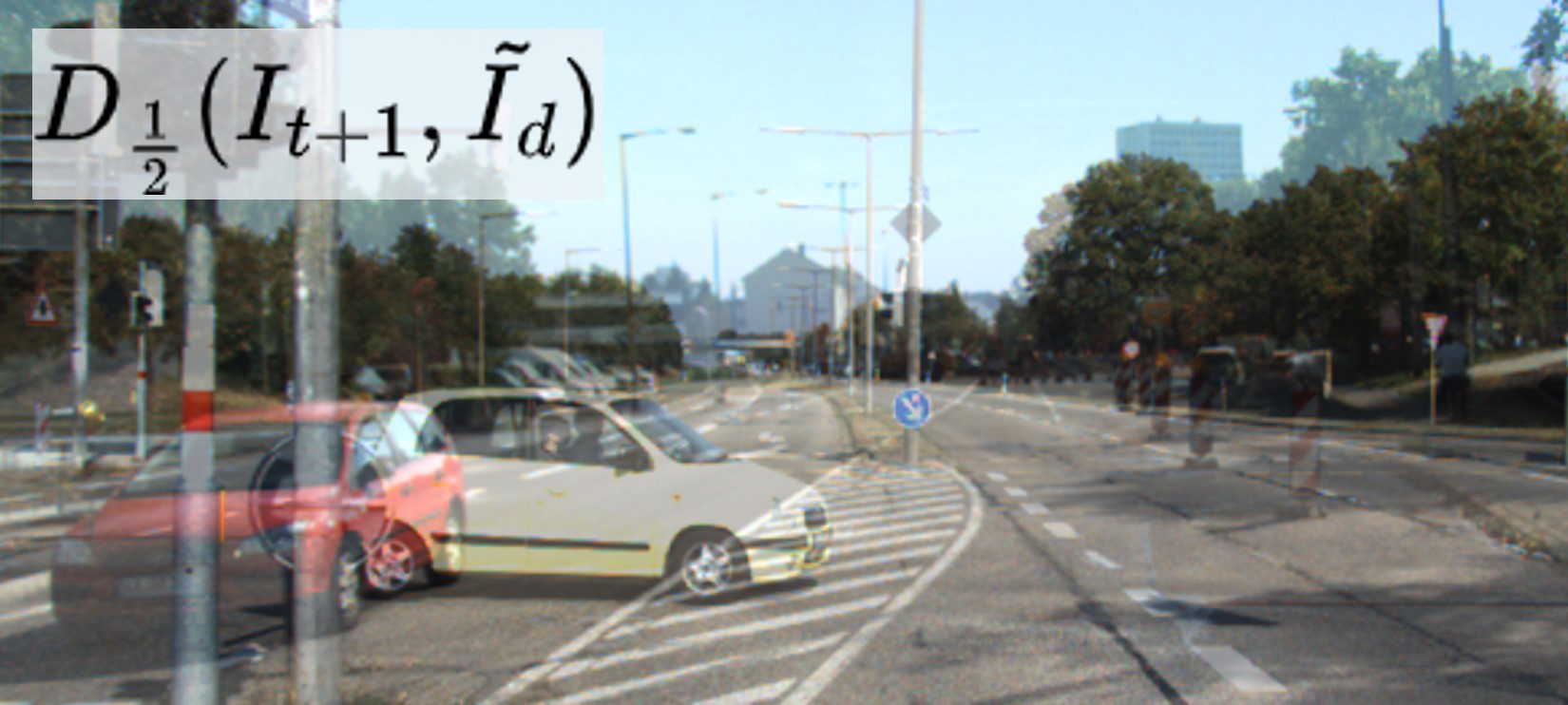}
& \hspace{-0.4cm}
\includegraphics[width=2.6cm]{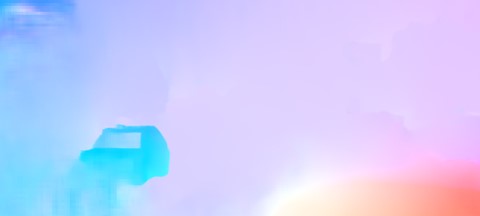} 
& \hspace{-0.5cm}
\includegraphics[width=2.6cm]{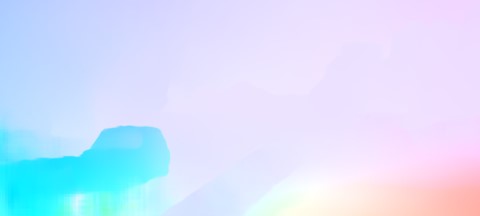}\\
\hspace{-0.2cm} 
\includegraphics[width=2.6cm]{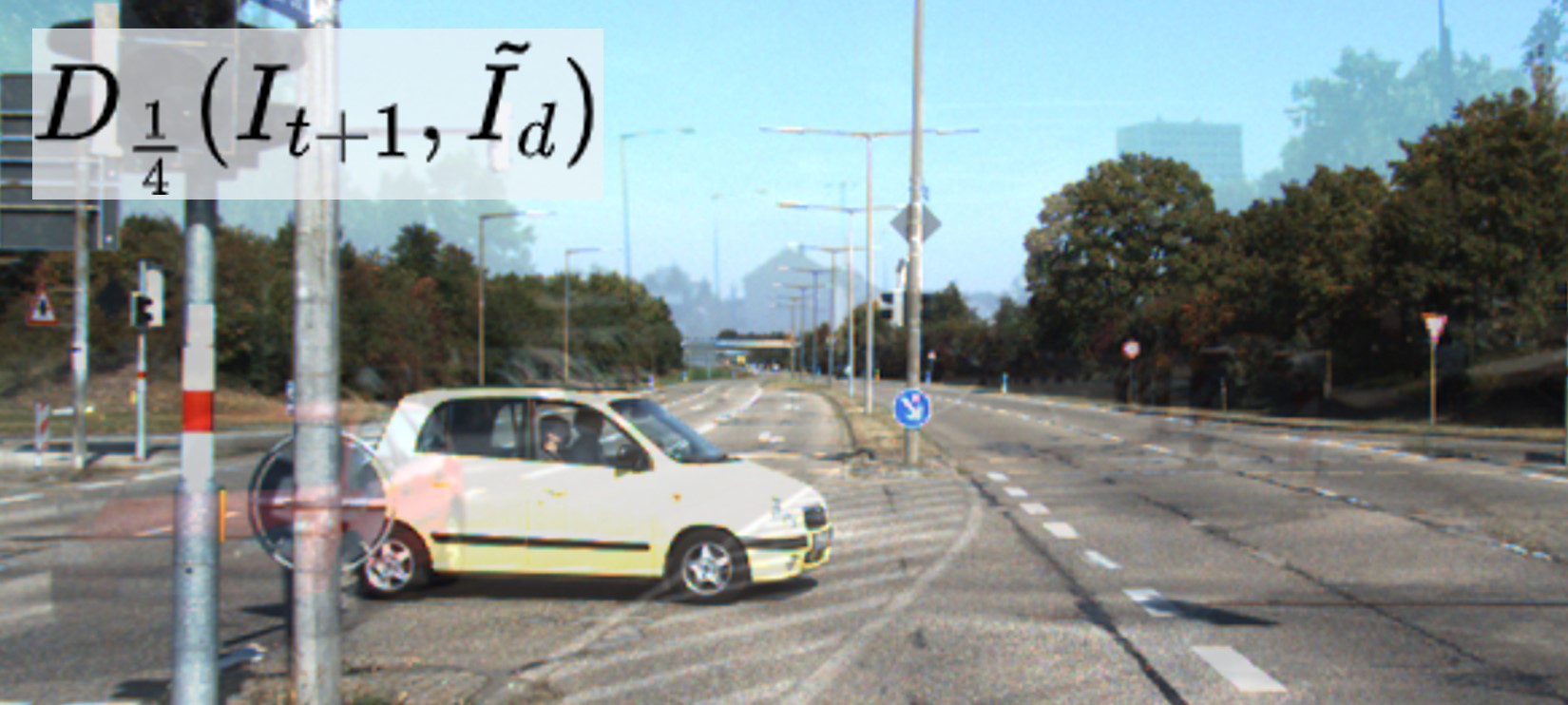}
& \hspace{-0.4cm}
\includegraphics[width=2.6cm]{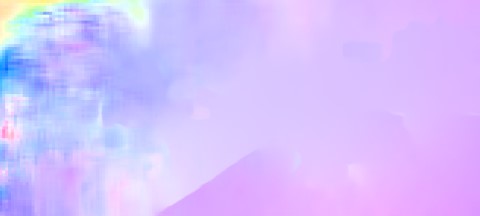} 
& \hspace{-0.4cm}
\includegraphics[width=2.6cm]{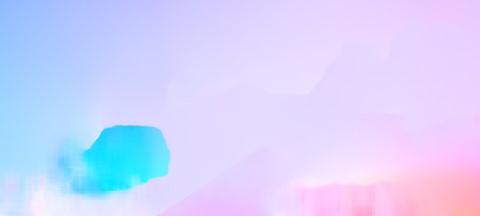}

\vspace{-7mm}
\end{tabular}$
\end{center}
% \vspace{-2mm}
\caption{Predictions on distracted video frame pairs. In the first column, the first row shows the overlaid original pair and the second to fourth rows show the distracted pairs. The mixing weight for the distractor increases from top to bottom. The RAFT model trained with DistractFlow performs robustly, while the original RAFT model completely fails when the distraction becomes large.
}
\vspace{-2mm}
\label{exp:qualitative}
\end{figure}

% XXX Let's move the following to supp XXX
% \textbf{Training Memory:} \fp{this seems not critical, can we shorten it to half?} \hc{[HC: Can we move this to supp? Seems more like an implementation detail rather than about the methodology.]} \fp{I agree!}
% For supervised training, our model requires twice the memory compared to the conventional model training. We additionally compute the loss for the output of $I_{1}$ and Distracted $I_{2}$. For semi-supervised training, we need additional memory to generate a forward-backward consistency-based confidence map. For the confidence map, we compute the output of ($I_{2}$,$I_{1}$) pair. However, such a large memory is required only for training, and it shows the same latency as an original model in the inference. 

\textbf{Robust Prediction:}
% \hc{[Can we do some feature visualization?]}
Figure \ref{exp:qualitative} shows the predictions of RAFT and DistractFlow-RAFT (trained with FlyingChairs and FlyingThings3D) on distracted frames pairs, using mixing ratios of 1, 0.75, 0.5, and 0.25. When the original $I_{t+1}$ has a small portion of a distracted image , the original RAFT has degraded performance. It completely fails to find the correspondence for the red car when $\lambda=0.25$. In contrast, the model trained using DistractFlow still robustly finds the correspondence in the distracted image.

% [It can be repeated objects problem that optical flow also has. But, in the similar scene (auto), the lane or signal or car's color or other backgrounds can be different. ]

% Fig.~\ref{exp:qualitative} shows the qualitative results for Original and Distracted image pairs using RAFT and DistractFlow (RAFT) models trained on C+T. There are 3 mixing $I_{2}$ images and all of them has correspondence with $I_{1}^{a}$. When image 2 has small portion ($\lambda \le 0.5$) of $I_{2}^{a}$, our model shows much better performance than origial RAFT model. 

\section{Conclusion}
\label{sec:conclusion}
\vspace{-3pt}
We proposed a novel method, DistractFlow, to augment optical flow training. We introduced realistic distractions to the video frame pairs which provided consistent improvements to optical flow estimation models. When unlabeled data was available, based on the original and distracted pairs, we devised a semi-supervised learning scheme using pseudo labels. We also incorporated forward-backward consistency through confidence maps that provided training stability and enhanced the performance further. Through extensive experiments on several optical flow estimation benchmarks: SlowFlow, Sintel, and KITTI, we showed that our method achieved significant improvements over the previous state of the art without inducing additional complexity during inference. In particular, models trained using our DistractFlow strategy are more robust in practical, challenging scenarios (e.g., consistent error reductions on SlowFlow despite strong motion blurs).
\newpage
%%%%%%%%% REFERENCES
{\small
\bibliographystyle{ieee_fullname}
\bibliography{egbib}
}

\clearpage
\newpage

\section{Implementation and Training Details}
\label{sec:intro}
We use the official codes for RAFT \cite{teed2020raft},\footnote{\url{https://github.com/princeton-vl/RAFT}} GMA\cite{jiang2021learning},\footnote{ \url{https://github.com/zacjiang/GMA}}  and FlowFormer\cite{huang2022flowformer}.\footnote{\url{https://github.com/drinkingcoder/FlowFormer-Official}}  We use 2 NVIDIA A100 GPUs for all training experiments. We follow RAFT and
GMA learning parameters except for the batch size. Specifically, we use a batch size of 8 for RAFT and GMA, when reproducing the baseline numbers and the numbers with DistractFlow. For FlowFormer, we follow their original setting including the batch size to reproduce this baseline. When applying DistractFlow, we use a batch size of 2 for FlowFormer due to GPU memory constraint.

\begin{table}[h]
\centering
\caption{EPE and coverage at varying confidence thresholds($\tau$). Test results obtained using RAFT on Sintel (final) dataset (trained on FlyingChair+FlyingThings3D).}
\label{tab:cal}
\vspace{-7pt}
\adjustbox{max width=0.45\textwidth}
{
\begin{tabular}{|c||c|c|c|c|c|c|c|}
\hline
Threshold ($\tau$) & 0.0 & 0.37 & 0.7 & 0.8 & 0.90 & 0.95 & 1.0 \\
\hline
EPE & 2.73 & 1.12 & 0.93 & 0.85 & 0.74 & 0.66 & 0.28\\
Coverage & 1.0 & 0.89 & 0.87 & 0.85 & 0.81 & 0.74 & 0.18$\times 10^{-4}$\\
% \% of confident area & 100 & 89.1 & 86.6 & 84.8 & 80.6 & 74.3 & 1.8$\times 10^{-3}$\\
\hline
\end{tabular}
}
% \vspace{-17pt}
\end{table}

\section{Calibration}
% \textbf{Confidence:} 
Table~\ref{tab:cal} shows that when choosing pixels with a higher confidence threshold, the EPE of the selected pixels decreases while the coverage (i.e., percentage of selected pixels) also decreases. This verifies that our confidence measure of Eq.~(5) in the paper properly captures the correctness of the prediction. Therefore, when applying a confidence threshold, we obtain a set of more accurately predicted pixels, rather than simply a smaller subset.

\begin{table}[h]
\begin{center}
\caption{Optical flow estimation results on Sintel (train/clean) dataset. The training is under the same as semi-supervised setting of Table 1 in the main paper.
}
\vspace{-3mm}
\label{sub:tab:supandsemi}
\adjustbox{max width=0.47\textwidth}
{
\begin{tabular}{|l|l||c|}
\hline

Method & Model & Sintel (Clean) \\
% \cline{3-6}
%  &  & (clean) & (Final) & (Fl-epe) & (Fl-all)\\
\hline
\hline

Supervised & \multirow{3}{*}{RAFT \cite{teed2020raft}} & 1.42 \\
FlowSupervisor \cite{im2022semi} & & 1.30 \\
DistractFlow (Our) && \textbf{1.25}\\
\hline
Supervised & \multirow{2}{*}{GMA \cite{jiang2021learning}} & 1.35\\
DistractFlow (Our) && \textbf{1.22}\\
\hline
Supervised & \multirow{2}{*}{FlowFormer \cite{huang2022flowformer}} & 0.95\\
DistractFlow (Our) && \textbf{0.90} \\
\hline
\end{tabular}
}
\vspace{-5mm}
\end{center}
\end{table}

\section{Sintel Clean Results}
As can be seen in the table~\ref{sub:tab:supandsemi}, our proposed DistractFlow also provides consistent improvements for state-of-the-art models on Sintel Clean (train), despite that this is an easier split with low-noise, synthetic images.
% \subsection{Additional Qualitative Results}

\end{document}